\documentclass[10pt, journal, compsoc]{IEEEtran}
\usepackage[cmex10]{amsmath}
\usepackage{times}
\usepackage{epsfig}
\usepackage{graphicx}
\usepackage{amssymb}
\usepackage{algorithmic}
\usepackage{algorithm}
\usepackage{color}
\usepackage{tikz}
\usepackage{mathrsfs}
\usepackage{bm}
\usepackage{multicol}
\usepackage{multirow}
\usepackage{cite}
\usepackage[section]{placeins}
\usepackage{url}
\usepackage{array}
\usepackage{subfig}

\usepackage{accents}
\newlength{\dhatheight}
\newcommand{\doublehat}[1]{%
    \settoheight{\dhatheight}{\ensuremath{\hat{#1}}}%
    \addtolength{\dhatheight}{-0.35ex}%
    \hat{\vphantom{\rule{1pt}{\dhatheight}}%
    \smash{\hat{#1}}}}

  \newcolumntype{L}{>{\centering\arraybackslash}m{8mm}}

\ifCLASSINFOpdf
\else
\fi
\begin{document}

\title{Discriminative Multiple Instance Hyperspectral Target Characterization}

\author{Alina~Zare,~\IEEEmembership{Senior Member,~IEEE,}
        Changzhe~Jiao,~\IEEEmembership{Student~Member,~IEEE,}
        and~Taylor~Glenn
\thanks{A. Zare is with the Department
of Electrical and Computer Engineering, University of Florida, Gainesville,
FL, 32611 USA e-mail: azare@ufl.edu.  C. Jiao is with the Department
of Electrical Engineering and Computer Science, University of Missouri, Columbia,
MO, 65211 USA.  T. Glenn is with Precision Silver LLC, Gainesville, FL 32611 USA}
\thanks{Manuscript received June 20, 2016, revised April 30, 2017, revised August 23, 2017, accepted September 12, 2017. }}

\markboth{Accepted to IEEE TPAMI}%
{Zare \MakeLowercase{\textit{et al.}}: Multiple Instance Hyperspectral Target Characterization}

\maketitle

\begin{abstract}
In this paper, two methods for \emph{discriminative multiple instance target characterization}, MI-SMF and MI-ACE, are presented.  MI-SMF and MI-ACE estimate a discriminative target signature from imprecisely-labeled and mixed training data.  In many applications, such as  sub-pixel target detection in remotely-sensed hyperspectral imagery, accurate pixel-level labels on training data is often unavailable and infeasible to obtain. Furthermore, since sub-pixel targets are smaller in size than the resolution of a single pixel, training data is comprised only of mixed data points (in which target training points are mixtures of responses from both target and non-target classes).  Results show improved, consistent performance over existing multiple instance concept learning methods on several hyperspectral sub-pixel target detection problems.  
\end{abstract}


\IEEEpeerreviewmaketitle

\section{Introduction}
\label{sec:intro}
\IEEEPARstart{H}{yperspectral} sub-pixel target detection is used for a huge variety applications including search and rescue \cite{4939406}, explosive residue detection \cite{849543}, food safety and quality monitoring \cite{mehl2004development}, chemical plume detection \cite{funk2001clustering,theiler2005characterizing},  biomedical applications \cite{martin2012hyperspectral}, landmine and explosive hazard detection \cite{4389068}, among many others.  The goal of sub-pixel target detection is to locate all instances of a target within a hyperspectral scene given a known target signature. However, in many applications, obtaining a target signature is often difficult or infeasible. In this paper, two methods for estimating a discriminative target signature from mixed training samples with imprecise labels are presented.  The proposed target characterization approaches, MI-SMF (Multiple Instance Spectral Matched Filter) and MI-ACE (Multiple Instance Adaptive Cosine Estimator) estimate target signatures that optimize the widely-used SMF and ACE sub-pixel target detector responses on a training data set with multiple-instance-style imprecise labels.

 Sub-pixel target detection is a challenging and important hyperspectral image analysis task. Numerous sub-pixel detectors have been proposed in the literature \cite{manolakis:2003,  Glenn:2013,  akhter:2015,kwon:2006,nasrabadi:2014,Broadwater:2007,Nasrabadi:2008,Bajorski:2012,Ziemann:2015}.  However, nearly all of these detectors rely on having an accurate target spectral signature in advance.  Much of the continued development of sub-pixel target detectors is driven by the lack of availability of effective target signatures for particular applications \cite{Manolakis:2007}.   Commonly, in hyperspectral analysis, target signatures are obtained from \emph{spectral libraries} comprised of spectral signatures collected in either controlled laboratory settings, using hand-held spectrometers, or pulled manually from a hyperspectral scene.  However, these methods for obtaining target signatures are often found to be ineffective.  For example, laboratory or hand-held spectrometer measured signatures do not account for atmospheric or environmental conditions and, thus, these signatures do not effectively translate to remotely-sensed hyperspectral imagery.  Signatures pulled from a scene (with environmental and atmospheric conditions similar to testing scenarios) are more likely to be effective, however, this approach requires that \emph{pure} spectral signatures from a target of interest can be accurately identified and extracted from a scene.  In the case of sub-pixel targets, pure pixels containing only target response do not exist. Furthermore, being able to accurately locate a target in a remotely-sensed scene is challenging. Often Global Positioning System (GPS) coordinates of targets are known, however, the precision of these coordinates are limited by the accuracy of the co-registration of the imagery to GPS coordinates and the accuracy of the GPS device used to collect those coordinates.   Thus, target GPS coordinates generally only provide an approximate location of a target in a scene.  Namely, a region or set of pixels containing the target can be identified but the specific pixel-level labeling cannot be accurately obtained. Furthermore, since the targets are subpixel, they cannot visibly seen in the imagery and, thus, training labels cannot even be manually created.  MI-SMF and MI-ACE address all of these challenges in obtaining target signatures. As opposed to modifying the target detector for improved performance, MI-SMF and MI-ACE estimate the target signature that improves SMF and ACE detection performance. Since MI-SMF and MI-ACE estimate discriminative target signatures from training data, it leverages the benefits of pulling a target signature from a hyperspectral scene but does not require pure pixel instances or accurate pixel-level labeling of target in the training data. Furthermore, the discriminative signatures estimated by MI-SMF and MI-ACE can be easily interpreted to understand what characteristics of the target class distinguish it from the background.  In other words, in addition to optimizing SMF and ACE performance, the resulting signatures estimated by MI-SMF and MI-ACE are interpretable and provide insight into what are the discriminative, salient features of the target. In the case of hyperspectral sub-pixel target detection, these discriminative, salient features are the spectral wavelengths and the spectral characteristics of the target that differ from the background. These spectral characteristics have physical meaning that can then be studied and understood once uncovered. 

 The majority of sub-pixel detection techniques are statistical methods in which the target and background signals are modeled as random variables distributed according to some respective underlying probability distribution \cite{Broadwater:2007,manolakis:2003,Eismann:2012}.  The detection problem can then be posed as a binary hypothesis test with two competing hypotheses: target absent ($\boldsymbol{H}_0$) or target present ($\boldsymbol{H}_1$) and a detector can be designed using the generalized likelihood ratio test (GLRT) approach \cite{Kay:1993}. The spectral matched filter (SMF) \cite{manolakis:2003,Matteoli:2014,Theiler:2006,Nasrabadi:2008,Kay:1993} and the adaptive coherence/cosine estimator (ACE) \cite{Kraut:1999, Kraut:2001, basener:2010clutter} are two such effective and extremely widely used sub-pixel detection algorithms.  The hypotheses used for the SMF are:
\begin{eqnarray}
&&\mathbf{H}_0: \mathbf{x}\sim\mathcal{N}\left(0, \Sigma_b \right)\nonumber \\
&&\mathbf{H}_1: \mathbf{x}\sim\mathcal{N}\left(a\mathbf{s}, \Sigma_b\right)
\label{eqn:smf_model}
\end{eqnarray}
where $\boldsymbol{\Sigma}_b$ is the background covariance and $\boldsymbol{s}$ is the known target signature which is scaled by a target abundance, $a$. The square-root of the GLRT for \eqref{eqn:smf_model} results in the following as the SMF detector: 
\begin{eqnarray}
D_{SMF}(\mathbf{x}, \mathbf{s})=\frac{\mathbf{s}^T\boldsymbol{\Sigma}^{-1}_b(\mathbf{x}-\boldsymbol{\mu}_b)}{\sqrt{\mathbf{s}^T\boldsymbol{\Sigma}^{-1}_b\mathbf{s}}}
\label{eqn:SMF}
\end{eqnarray}
 where $\boldsymbol{\mu}_b$ is the background mean subtracted from the data to ensure a zero-mean background as defined in $\boldsymbol{H}_0$.  

In comparison, the hypotheses used for the unstructured-background ACE detector are:
\begin{eqnarray}
&&\mathbf{H}_0: \mathbf{x}\sim\mathcal{N}\left(0, \sigma_0^2\Sigma_b \right)\nonumber \\
&&\mathbf{H}_1: \mathbf{x}\sim\mathcal{N}\left(a\mathbf{s}, \sigma_1^2\Sigma_b\right)
\label{eqn:ace_model}
\end{eqnarray}
 which includes $\sigma_0^2 = \frac{1}{d}\mathbf{x}^T\Sigma_b^{-1}\mathbf{x}$ and $\sigma_1^2 = \frac{1}{d}\left( \mathbf{x} - a\mathbf{s}\right)^T\Sigma_b^{-1}\left( \mathbf{x} - a\mathbf{s}\right)$ to add scale-invariance to the ACE detector where $d$ is the dimensionality of the spectra.   The square-root of the GLRT for \eqref{eqn:ace_model} results in the following as the ACE detector \cite{Broadwater:2007,Kraut:2001,Kraut:1999}: 
\begin{eqnarray}
D_{ACE}(\mathbf{x}, \mathbf{s})=\frac{\mathbf{s}^T\boldsymbol{\Sigma}^{-1}_b(\mathbf{x}-\boldsymbol{\mu}_b)}{\sqrt{\mathbf{s}^T\boldsymbol{\Sigma}^{-1}_b\mathbf{s}}\sqrt{(\mathbf{x}-\boldsymbol{\mu}_b)^T\boldsymbol{\Sigma}^{-1}_b(\mathbf{x}-\boldsymbol{\mu}_b)}}.
\label{eqn:ACE}
\end{eqnarray}
Comparing \eqref{eqn:SMF} and \eqref{eqn:ACE}, we see that the difference between these two detectors is a normalization of an input test point.  As a result of this difference, the SMF detection statistics is a projection of an (unnormalized) test data point onto a target vector in a whitened coordinate space.  Since the test point is not normalized, data points with larger magnitude (of components not orthogonal to the target signature) result in larger detection statistics (i.e., in SMF, magnitude matters).  In contrast, ACE normalizes all input test points such that detection statistics are determined only by the vector angle between a test point and the target signature in the whitened coordinate space (and magnitude does not play a role).  
 In order to apply SMF or ACE, the target signature, $\mathbf{s}$, must be known.  The proposed MI-SMF and MI-ACE estimate a discriminative $\mathbf{s}$  from imprecisely-labeled, mixed training data that optimizes the SMF and ACE detection statistics.

\section{Related Methods}
The proposed problem of target characterization from imprecise labels is most closely related to multiple instance concept learning since, in those methods, a positive-class concept (i.e., a target signature) is also estimated from imprecisely-labeled training data. Here, a class concept refers to a generalized class prototype in the feature space.  Among the rapidly growing body of Multiple Instance Learning (MIL) methods \cite{Dietterich:1997,bolton:2011,Chen:2006}, only a few MIL methods estimate class concepts.  Most notably, the Diverse Density (DD) \cite{Maron:1998}, the Expectation-Maximization DD (EM-DD) \cite{Zhang:2002}, the Dictionary-based Multiple Instance Learning (DMIL) \cite{Shrivastava:2014,Shrivastava:2015} and the extended FUnctions of Multiple Instances ($e$FUMI) \cite{Zare:2010icpr,Jiao:2015} methods are MIL methods that estimate class concepts.  

Multiple instance learning is a variation on supervised learning for problems with imprecisely-labeled training data.  Instead of pairing each training point with a class label, MIL methods learn from a set of labeled ``bags'' in which a bag is defined to be a multi-set of data points.  Each bag is labeled as either a ``positive'' or ``negative'' bag.  A bag is defined to be positive if at least one of the data points in the bag is an instance of the positive target class.  The number of positive instances in each positive bag is unknown.  Negative bags are composed entirely of non-target data points.  An advantage of the MIL concept learning methods is that concepts can then be examined after applying the MIL approach to obtain insight into what characterizes the target class.  In the case of hyperspectral sub-pixel target detection, this is extremely useful as the discriminative spectral characteristics in particular spectral wavelengths have physical meaning.  By uncovering the discriminative spectral characteristics, the physical properties of the target material that result in these  characteristics can be uncovered and studied. 

Diverse density \cite {Maron:1998} finds a positive-class concept  that lies close to at least one instance in each positive bag and maximizes the  distance from all instances in negatively labeled bags. The distance measure used by DD to determine how close the concept is to the instances in positive bags and how far it is from the instances in negative bags is Euclidean distance.  Namely, DD estimates the positive class concept, $\mathbf{s}$, that maximizes the following Noisy-OR objective:
\begin{eqnarray}
& &\arg\max_{\mathbf{s}}\prod_{i} Pr\left( \mathbf{s}  | B_i^+\right) \prod_{i } Pr\left( \mathbf{s}  | B_i^-\right)\nonumber\\
& &\quad=\prod_{i} 1 - \prod_{j} 1- \exp\left\{-\left\| \mathbf{B}^+_{ij} - \mathbf{s}\right\|^2\right\}\label{eqn:dd}\\
&  & \quad\quad \prod_{i}\prod_{j}  1 - \exp\left\{-\left\| \mathbf{B}^-_{ij} - \mathbf{s}\right\|^2\right\} \nonumber
\end{eqnarray}
where $B_{ij}^+$ is the $j^{th}$ point in the $i^{th}$ positive bag and $B_{ij}^-$ is the $j^{th}$ point in the $i^{th}$ negative bag.  

EM-DD, the Expectation-Maximization (EM) version of diverse density \cite{Zhang:2002}, estimates a target concept using an EM approach in which, during the $E$-step, a single instance from each bag is selected as the one most likely to be cause of the bag's label (e.g., for positive bags, the selected instance is the instance mostly likely to be the positive example in the bag). Then, during the $M$-step, the concept is updated using gradient ascent.  Zhang and Goldman \cite{Zhang:2002} argue that EM-DD improves accuracy and computation time over the DD algorithm since the use of a single instance from each bag simplifies the search space and helps to avoid getting caught in local minima (by encouraging large jumps when the selected instances are changed in a bag each iteration).  

DMIL \cite{Shrivastava:2014,Shrivastava:2015}, instead of learning a single class concept close to the conjunction of positive bags and far from each negative instance, estimates class-specific dictionaries (one for each class) by enforcing that at least one instance in each positive bag for a class is well represented by the class-specific dictionary and all negative instances are poorly represented by that dictionary.  The dictionaries are estimated by maximizing the Noisy-OR model in \eqref{eqn:dd} where, instead of using the Euclidean distance to measure the dissimilarity between each instance and the associated class concept, the reconstruction error of an instance using the class-specific dictionary is used.

$e$FUMI \cite{Jiao:2015}, like DMIL, estimates a full dictionary as opposed to a single concept.  In contrast, however, $e$FUMI does not estimate distinct class-specific dictionaries.  A single dictionary with one target concept and a shared non-target concept dictionary is estimated.  Each instance is modeled as a convex combination of positive and/or negative concepts and estimates the target and non-target concepts using an EM approach in which the hidden latent variable are the labels for each instance in the training data set.  

MI-SMF and MI-ACE, like DD and EM-DD, estimate a target concept.  However, instead of using a Euclidean distance to measure the similarity between instances and the target concept, MI-SMF and MI-ACE use the cosine similarity which, as shown in the following section, is closely aligned with the SMF and ACE target detectors.  The cosine similarity is found to be more robust in the case of mixed training data in which target signatures are sub-pixel components of positive training data points.  

\section{MIL Target Characterization}
Let $\mathbf{X}=\left[\mathbf{x}_1,\cdots,\mathbf{x}_N\right]\in\mathbb{R}^{d\times N}$ be training data where $d$ is the dimensionality of an instance, $\mathbf{x}_i$, and $N$ is the total number of training instances. The data is grouped into $K$ \textit{bags},  $\mathbf{B} = \left\{ \mathbf{B}_1, \ldots, \mathbf{B}_K\right\}$, with associated binary bag-level labels, $L = \left\{L_1, \ldots, L_K\right\}$ where $L_j \in \left\{ 0, 1\right\}$ and $\mathbf{x}_{ji} \in \mathbf{B}_j$ denotes the $i^{th}$ instance in bag $\mathbf{B}_j$.  Positive bags (i.e., $\mathbf{B}_j$ with $L_j = 1$, denoted as $\mathbf{B}_j^+$) contain at least one instance composed of some target:
\begin{eqnarray} \label{eq:l1}
&\text{if } L_j = 1,&   \exists \mathbf{x}_{ji} \in \mathbf{B}_j^+ \\
&&\text{ s.t. } \mathbf{x}_{ji} \sim \mathcal{N}\left( \alpha_{it}\mathbf{s} + \boldsymbol{\mu}_b, \sigma_1^2\Sigma_b \right) , \alpha_{it} \ne 0 \nonumber
\end{eqnarray}
However, the number of instances in a positive bag with a target component is unknown.  If $\mathbf{B}_j$ is a negative bag (i.e., $L_j = 0$, denoted as $\mathbf{B}_j^-$), then this indicates that $\mathbf{B}_j^-$ does not contain any target:
\begin{equation}
\text{if }L_j = 0,   \mathbf{x}_{ji} \sim \mathcal{N}\left( \boldsymbol{\mu}_b, \sigma_0^2\Sigma_b \right) \forall \mathbf{x}_{ji} \in \mathbf{B}_j^-
 \label{eq:l2}
\end{equation}

Given this problem formulation, the goal of MI-SMF and MI-ACE is to estimate the target signature, $\mathbf{s}$, that maximizes the corresponding detection statistic for the target instances in each positive bag and minimize the detection statistic over all negative instances.  This is accomplished by maximizing the following objective: 
\begin{equation}
\arg\max_{\mathbf{s}} \frac{1}{N^+} \sum_{j: L_j = 1} D(\mathbf{x}_j^{\ast}, \mathbf{s}) - \frac{1}{N^-}\sum_{j:L_j = 0}\frac{1}{N_j^-}\sum_{\mathbf{x}_i \in B_j^-} D(\mathbf{x}_i, \mathbf{s})
\label{eqn:objective}
\end{equation}
where $N^+$ and $N^-$ are the number of positive and negative bags, respectively, $N_j^-$ is the number of instances in the $j^{th}$ negative bag, and $\mathbf{x}_j^{\ast}$ is the selected instance from the positive bag $B_j^+$ that is mostly likely a target instance in the bag. The selected instance is identified as the point with the maximum detection statistic given a target signature, $\mathbf{s}$: 
\begin{equation}
\mathbf{x}_j^{\ast} = \arg\max_{\mathbf{x}_i \in B_j^+}  D(\mathbf{x}_i , \mathbf{s}) 
\label{eqn:selected1}
\end{equation}
The use of the selected instance allows MI-SMF and MI-ACE to inherit the advantages of doing so outlined in the EM-DD paper \cite{Zhang:2002}.  

Given a set of selected instances, the target signature can be estimated by maximizing \eqref{eqn:objective} with respect to $\mathbf{s}$. Let us first consider ACE and MI-ACE. To derive the update equation for the target signature, first note that the ACE detector can be re-written as follows: 
\begin{align}
D&_{ACE}(\mathbf{x}, \mathbf{s})= \frac{\mathbf{s}^T\boldsymbol{\Sigma}^{-1}_b(\mathbf{x}-\boldsymbol{\mu}_b)}{\sqrt{\mathbf{s}^T\boldsymbol{\Sigma}^{-1}_b\mathbf{s}}\sqrt{(\mathbf{x}-\boldsymbol{\mu}_b)^T\boldsymbol{\Sigma}^{-1}_b(\mathbf{x}-\boldsymbol{\mu}_b)}}\\
=&\frac{\mathbf{s}^T\mathbf{U}\boldsymbol{D}^{-\frac{1}{2}}\boldsymbol{D}^{-\frac{1}{2}}\mathbf{U}^T(\mathbf{x}-\boldsymbol{\mu}_b)}{\sqrt{\mathbf{s}^T\mathbf{U}\boldsymbol{D}^{-\frac{1}{2}}\boldsymbol{D}^{-\frac{1}{2}}\mathbf{U}^T\mathbf{s}}\sqrt{(\mathbf{x}-\boldsymbol{\mu}_b)^T\mathbf{U}\boldsymbol{D}^{-\frac{1}{2}}\boldsymbol{D}^{-\frac{1}{2}}\mathbf{U}^T(\mathbf{x}-\boldsymbol{\mu}_b)}}\\
=& \left( \frac{\hat{\mathbf{s}}}{\left\|\hat{\mathbf{s}}\right\|}   \right)^T\left( \frac{\hat{\mathbf{x}}}{\left\|\hat{\mathbf{x}}\right\|}   \right)\\
=& \doublehat{\mathbf{s}}^T\doublehat{\mathbf{x}}
\end{align}
where $\hat{\mathbf{x}} = \mathbf{D}^{-\frac{1}{2}}\mathbf{U}^T(\mathbf{x}-\boldsymbol{\mu}_b)$,  $\hat{\mathbf{s}} = \mathbf{D}^{-\frac{1}{2}}\mathbf{U}^T\mathbf{s}$, $\mathbf{U}$ and $\mathbf{D}$ are the eigenvectors and eigenvalues of the background covariance matrix, $\boldsymbol{\Sigma_b}$, respectively, $\doublehat{\mathbf{s}} = \frac{\hat{\mathbf{s}}}{\left\|\hat{\mathbf{s}}\right\|} $  and $\doublehat{\mathbf{x}} = \frac{\hat{\mathbf{x}}}{\left\|\hat{\mathbf{x}}\right\|} $.  Here, it can be clearly noted that the ACE detection statistic is the cosine similarity between a test data point, $\mathbf{x}$, and a target signature, $\mathbf{s}$, in a whitened coordinate space.  Thus, the objective function in \eqref{eqn:objective} can be rewritten for MI-ACE as: 
\begin{align}
\arg&\max_{\doublehat{\mathbf{s}}} \frac{1}{N^+} \sum_{j: L_j = 1} \doublehat{\mathbf{s}}^T\doublehat{\mathbf{x}}^{\ast}_j - \frac{1}{N^-}\sum_{j:L_j = 0}\frac{1}{N_j^-}\sum_{\mathbf{x}_i \in B_j^-} \doublehat{\mathbf{s}}^T\doublehat{\mathbf{x}_i} \nonumber \\
& \text{ such that } \doublehat{\mathbf{s}}^T\doublehat{\mathbf{s}} =1.
\label{eqn:objective2}
\end{align}
The constraint, $\doublehat{\mathbf{s}}^T\doublehat{\mathbf{s}} =1$, is a result of the fact that $\doublehat{\mathbf{s}} = \frac{\hat{\mathbf{s}}}{\left\|\hat{\mathbf{s}}\right\|}$ and aids in preventing values of $\doublehat{\mathbf{s}}$ from being arbitrarily large to maximize the first term in \eqref{eqn:objective2}.  Now, given \eqref{eqn:objective2}, the update equation for $\doublehat{\mathbf{s}}$ can derived by solving the associated Lagrangian resulting in:
\begin{equation}
\doublehat{\mathbf{s}} = \frac{\mathbf{t}}{\left\|\mathbf{t}\right\|} \text{ where } \mathbf{t} = \frac{1}{N^+} \sum_{j: L_j = 1} \doublehat{\mathbf{x}}^{\ast}_j - \frac{1}{N^-}\sum_{j:L_j = 0}\frac{1}{N_j^-}\sum_{\mathbf{x}_i \in B_j^-} \doublehat{\mathbf{x}_i} 
\label{eqn:tupdate}
\end{equation}

MI-SMF can be similarly derived.  As noted in Section \ref{sec:intro}, MI-SMF does not normalize input test points.  Thus, the difference between MI-SMF over MI-ACE is the use of $\hat{\mathbf{x}}$ instead of $\doublehat{\mathbf{x}}$ in the objective function and target signature update equation.  For MI-SMF, the objective function can be written as:
\begin{align}
\arg&\max_{\doublehat{\mathbf{s}}} \frac{1}{N^+} \sum_{j: L_j = 1} \doublehat{\mathbf{s}}^T\hat{\mathbf{x}}^{\ast}_j - \frac{1}{N^-}\sum_{j:L_j = 0}\frac{1}{N_j^-}\sum_{\mathbf{x}_i \in B_j^-} \doublehat{\mathbf{s}}^T\hat{\mathbf{x}_i} \nonumber\\
&\text{ such that } \doublehat{\mathbf{s}}^T\doublehat{\mathbf{s}} =1.
\label{eqn:objective3}
\end{align}
resulting in the following update equation for $\doublehat{\mathbf{s}}$:
\begin{equation}
\doublehat{\mathbf{s}} = \frac{\mathbf{t}}{\left\|\mathbf{t}\right\|} \text{ where } \mathbf{t} = \frac{1}{N^+} \sum_{j: L_j = 1} \hat{\mathbf{x}}^{\ast}_j - \frac{1}{N^-}\sum_{j:L_j = 0}\frac{1}{N_j^-}\sum_{\mathbf{x}_i \in B_j^-} \hat{\mathbf{x}_i}.
\label{eqn:tupdate2}
\end{equation}

The update for $\doublehat{\mathbf{s}} $ in MI-SMF and MI-ACE has a closed form solution, so, unlike many MIL methods a gradient ascent approach is not needed when updating the target concept. Also, note that the second term in \eqref{eqn:tupdate} and \eqref{eqn:tupdate2} does not change for the life of the algorithm and can be precomputed. 

 MI-SMF and MI-ACE proceed by alternating between selecting representative instances for each positive bag and updating the target concept.  The resulting methods, summarized in Alg. \ref{alg:miace}\footnote{Our MI-SMF and MI-ACE implementations are available: https://github.com/GatorSense/}, are straight-forward, fast, and effective approaches for multiple instance target characterization. 

\begin{algorithm}
\caption{MI-SMF/MI-ACE}
\algsetup{indent=2em}
\begin{algorithmic}[1] 
\STATE Compute $\boldsymbol{\mu_b}$ and $\boldsymbol{\Sigma}_b$ as the mean and covariance of all instances in the negative bags
\STATE Subtract the background mean and whiten all instances, $\hat{\mathbf{x}} = \mathbf{D}^{-\frac{1}{2}}\mathbf{U}^T(\mathbf{x}-\boldsymbol{\mu}_b)$
\STATE If MI-ACE, normalize: $\doublehat{\mathbf{x}} = \frac{\hat{\mathbf{x}}}{\left\|\hat{\mathbf{x}}\right\|}$
\STATE Initialize $\doublehat{\mathbf{s}}$ using the instance in a positive bag resulting in largest objective function value
\REPEAT
\STATE Update the selected instances, $\mathbf{x}_j^{\ast}$, for each positive bag, $\mathbf{B}_j^+$ using  \eqref{eqn:selected1} 
\STATE Update $\doublehat{\mathbf{s}}$ using \eqref{eqn:tupdate} for MI-ACE or \eqref{eqn:tupdate2} for MI-SMF
\UNTIL{Stopping Criterion Reached}
   \RETURN $\mathbf{s} = \frac{\mathbf{t}}{\left\| \mathbf{t}\right\|}$ where $\mathbf{t} = \mathbf{U}\mathbf{D}^{\frac{1}{2}}\doublehat{\mathbf{s}}$
\end{algorithmic} 
\label{alg:miace}
\end{algorithm}

Note that for each set of selected instances, the $\doublehat{\mathbf{s}}$ is determined using a closed-form update.  Thus, given the same set of selected instances, the same $\doublehat{\mathbf{s}}$ will be calculated.  Given that there are a finite set of possible selected instances, there are a finite set of $\doublehat{\mathbf{s}}$ estimates.  MI-SMF and MI-ACE terminates when $\doublehat{\mathbf{s}}$ is repeated indicating that the same set of selected instances were chosen; this can occur in contiguous iterations or not.  This convergence sketch mimics that of the one described in \cite{Zhang:2002}.  In practice we found that MI-SMF and MI-ACE generally converged to a solution in less than 7 iterations. 

It can also be noted that the ACE target detector is a non-linear detector in the original spectral space but a linear discriminant in the whitened, normalized coordinate space.  Thus, a multiple instance linear discriminant can be estimated using a procedure similar to Alg. \ref{alg:miace} by eliminating steps (2)-(3) in the method (and, to estimate a bias, appending a value of 1 to each input test point).  Thus, Alg. 1 can be applied to any data as an approach to estimate a linear discriminant from data burdened with uncertain labels. 

\section{Experimental Results}
In the following MI-SMF and MI-ACE are evaluated and compared to several MIL concept learning methods on simulated data and to a real hyperspectral target detection data set.  The simulated data experiments are included to illustrate the properties of MI-SMF and MI-ACE and provide insight into how and when the methods are effective.  
\subsection{Simulated Data}
Simulated data sets were generated following the hyperspectral linear mixing model \cite{zare2014endmember} using the approach outlined in Alg. 3 and 4 of \cite{Jiao:2015}.  Namely, for each instance in a negative bag and negative instances in positive bags, a uniform random number of non-target signatures were selected and the selected non-target instances were combined to generate the instance using a convex combination with proportions drawn from a uniform Dirichlet distribution.  Similarly, for each true positive instance, a uniform random number of non-target signatures were selected and the target signature along with selected non-target instances were combined to generate the instance using a convex combination with proportions drawn from a Dirichlet distribution. For positive instances, the $\boldsymbol{\alpha}$ parameters of the Dirichlet were set to achieve the desired level of sub-pixel target mixing and variance in mixing proportions.\footnote{Simulated data generation code is available: \url{https://github.com/GatorSense/FUMI/tree/master/gen_synthetic_data_code}}
\subsubsection{Estimated Discriminative Target Concept vs. True Target Signature}
This first experiment is to illustrate that the \emph{discriminative} target concept estimated by MI-SMF and MI-ACE is not necessarily equal to the \emph{true} underlying target signature. In this experiment, two simulated two-dimensional data sets (for easy visualization) were generated.  The data was generated with two background and one target endmember (i.e., material signature), 10 positive and 10 negative bags, each bag contained 10 instances with only 3 target instances in each positive bag.  The target data points had 0.2 proportion of target on average. For the first data set, target points are randomly mixed with either or both of the background materials.  For the second data set, the target is only mixed with one of background materials (e.g., thus, simulating the case that targets only appear in certain context or around certain materials).  Zero-mean Gaussian noise was added such that the data has an SNR of 20dB.  

\begin{figure}[!h]
\centering
\subfloat[Simulated 2D Data 1: True Target Signature and estimated MI-SMF and MI-ACE Target Concepts, Target data points are shown as stars.  Color of data points indicate bag number. ]{\includegraphics[trim={2.1cm 6.5cm 2.2cm 7cm},clip,width=2.5in]{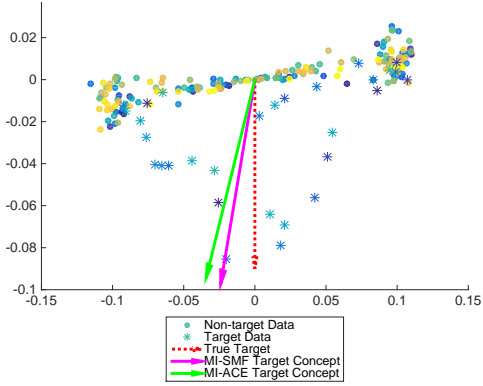}%
\label{fig:simexp1_1}}

\subfloat[Simulated 2D Data 2: True Target Signature and estimated MI-SMF and MI-ACE Target Concepts, Target data points are shown as stars.  Color of data points indicate bag number. ]{\includegraphics[trim={2.3cm 6.6cm 2.2cm 7cm},clip,width=2.5in]{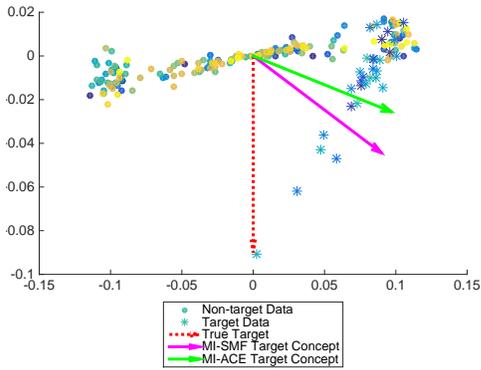}%
\label{fig:simexp1_2}}
\caption{MI-SMF and MI-ACE estimated target concepts in comparison to the true target signature used to simulate the data.  MI-SMF and MI-ACE do not necessarily recover the true target signature as shown by Simulated 2D Data Set 2.  Instead, the methods estimate the signature that maximizes detection performance. }
\label{fig_sim}
\end{figure}

\begin{figure}[!h]
\centering

\subfloat[Simulated 2D Data 1 ROC Curves: ROC curves for True Target Signature using SMF and ACE, MI-SMF and MI-ACE ]{\includegraphics[trim={2cm 7cm 2.3cm 7cm},clip,width=2.5in]{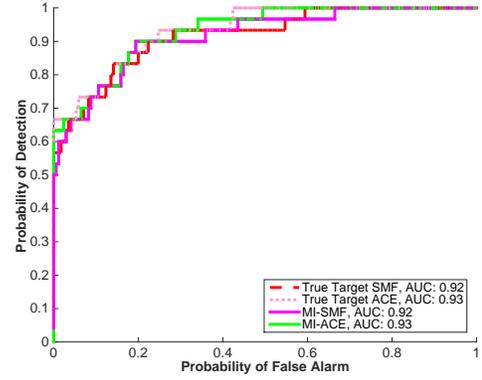}%
\label{fig:simexp1_3}}

\subfloat[Simulated 2D Data 2 ROC Curves: ROC curves for True Target Signature using SMF and ACE, MI-SMF and MI-ACE]{\includegraphics[trim={2cm 7cm 3cm 7cm},clip,width=2in]{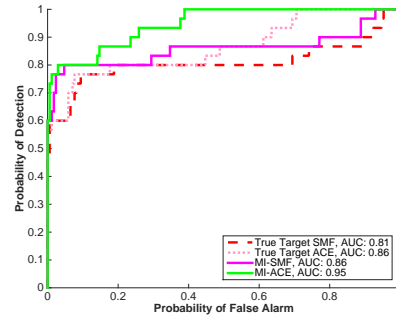}%
\label{fig:simexp1_4}}

\caption{ ROC curve analysis using SMF and ACE with the true target signature in comparison to MI-SMF and MI-ACE estimated target concepts. MI-SMF and MI-ACE do not necessarily recover the true target signature as shown by Simulated 2D Data Set 2.  Instead, the methods estimate the signature that maximizes detection performance. }
\label{fig_sim_roc}
\end{figure}

In this experiment, when training MI-ACE, the global mean and covariance over both positive and negative bags are used during whitening (\emph{i.e.}, the global mean is the mean over all training data points across both positive and negative bags and the global covariance is the covariance matrix computed over all training data points across both positive and negative bags).  This is done since, in the case of low-dimensional data, the normalization step using only the negative bag mean and covariance corrupt the structure of the data (note: this is not the case in high-dimensional hyperspectral data).  

 The true target vector used to generate the data, all data samples and bags, and the estimated discriminative target concepts using MI-SMF and MI-ACE are shown in Fig. \ref{fig:simexp1_1} and \ref{fig:simexp1_2}.  Using the estimated target concepts and the true target vector, the SMF and ACE detectors were applied to the data and the resulting ROC (receiver operating characteristic) curves  are shown in Fig. \ref{fig:simexp1_3}  and \ref{fig:simexp1_4}.  For the true target signatures, before applying SMF or ACE, the background mean is subtracted from the true target signature (as it improves performance of the detectors). For the first simulated 2D data set, both MI-SMF and MI-ACE estimate a signature very similar to the true target signatures.  However, for the second data set, neither MI-SMF or MI-ACE recover the true target signature from the data but instead estimate a target concept that maximizes target detection performance.  In the second data set,  the target is highly mixed with only one of the background endmembers and this additional contextual information  is learned during MI-SMF and MI-ACE training and leveraged.  For the second simulated data set, the area under the ROC curves (AUC) for the true target signature using the SMF and ACE detectors were 0.81 and 0.86, respectively.  In contrast, the AUC for the MI-SMF and MI-ACE target concepts using the SMF and ACE detectors, respectively, were 0.86 and 0.95.  

\subsubsection{Simulated Hyperspectral Data}
\begin{figure}[!ht]
\centering
\includegraphics[trim={0cm 5.6cm 0cm 6cm},clip,width=3in]{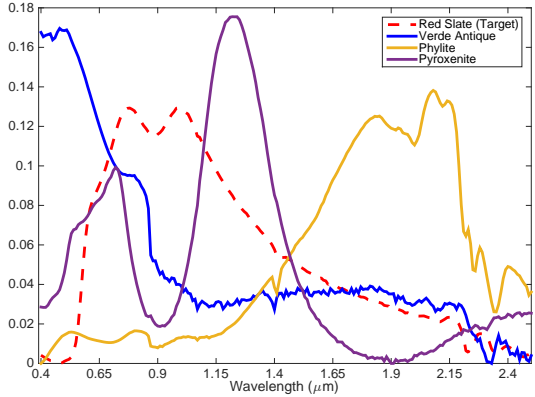}
\caption{Hyperspectral Signatures used to generate hyperspectral simulated data set. Signatures selected from the ASTER spectral library include Red Slate (target), Verde Antique, Phyllite and Pyroxenite spectra. }
\label{fig:constituent_endmembers}
\end{figure}

In the second set of simulated data experiments, a hyperspectral data set was simulated based on the linear mixing model using one target and three background spectra selected from the ASTER spectral library \cite{aster:2009}. Specifically, the Red Slate, Verde Antique, Phyllite and Pyroxenite spectra from the rock class with 211 bands and wavelengths ranging from $0.4 \mu$m to $2.5 \mu$m (as shown in Fig. \ref{fig:constituent_endmembers}) were used as endmembers to generate hyperspectral data. Red Slate was labeled as the target endmember.   Results of MI-ACE and MI-SMF were compared to EM-DD (estimating both a point and scale value) \cite{rahmani2005localized,Zhang:2002} and $e$FUMI \cite{Jiao:2015} such that the estimated target concepts can be compared.  In all of these experiments, separate training and testing data sets were generated and zero-mean Gaussian noise was added to the simulated training and testing data such that the SNR was 20dB.  All of the testing data sets were generated with 25,000 true negative and  25,000 true positive points with an average target proportion value of 0.15.  The parameters for generating the testing data were held constant such that results obtained using different training sets could be directly compared.  In all of these experiments, the target concept estimated from the training data by MI-SMF, MI-ACE and $e$FUMI, were evaluated using the SMF detection statistic for MI-SMF and the ACE detection statistic for MI-ACE and $e$FUMI on the test data.  The target point and scaling values estimated by EM-DD were evaluated on test data using the prediction approach outlined by Zhang and Goldman \cite{Zhang:2002}.  Namely, for each test data point, the detection statistic is computed using \eqref{eqn:EM-DDprd}:
\begin{equation}
Pr(\mathbf{x}_i \in \mathbf{s}) = \exp\left\{ - \sum_{d=1}^D {s}^c_{d} \left({x}_{id} - {s}^p_{d}\right)^2 \right\}
\label{eqn:EM-DDprd}
\end{equation}
where $\mathbf{x}_i$ is the $i^{th}$ test data point, ${x}_{id}$ is the $d^{th}$ feature value of test point $i$, $D$ is the dimensionality of the data, ${s}^c_{d}$ is the estimated EM-DD scaling value for the $d^{th}$ dimension and ${s}^p_{d}$ is the EM-DD point value for the $d^{th}$ dimension. $e$FUMI was initialized as outlined in \cite{Jiao:2015} and run with parameter settings of $\mu = 0.05, \alpha = 1.2, \beta=60, M=7$, and $\Gamma = 10$. These $e$FUMI parameters were determined manually to maximize $e$FUMI performance. As outlined in \cite{Jiao:2015}, non-target signatures were initialized using the VCA algorithm \cite{nascimento:2005} on all data in the negatively-labeled bags. Then, using these initial non-target signatures, the data point with the largest reconstruction error when representing each data point as a linear combination of initial non-target signatures is set as the initial target signature, $\mathbf{e}_T$. EM-DD scaling values were all initialized to one and the target point value was initialized to the same initial positive data point as used by $e$FUMI. 

\paragraph{Varying Number of Positive vs. Negative Bags}  In the first experiment with the simulated hyperspectral data, the number of positive vs. negative bags was varied to investigate if there is any sensitivity of the MI-ACE and MI-SMF methods to the proportion of positive bags in the training data. In this experiment, the total number of bags in the training data was held constant at 50 with the proportion of positive bags being varied from 0.25, 0.15 to 0.05 (corresponding to 13, 8, and 3 positive bags, respectively).  Each bag contained ten data points with positive bags containing only 2 true target points with an average target proportion of 0.05.  This resulted in a very highly mixed data set.  

\begin{figure}[!ht]
\centering
\includegraphics[trim={1cm 6cm 1cm 6cm},clip,width=3in]{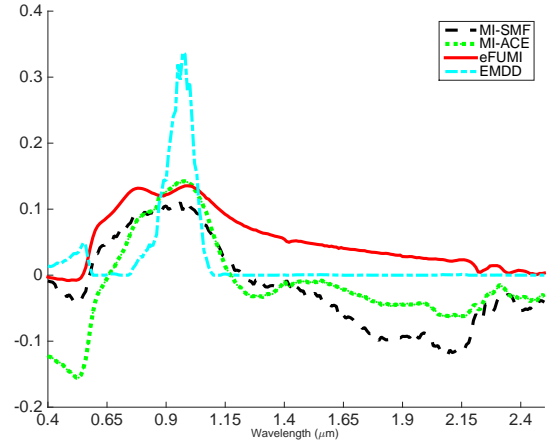}%
\caption{MI-SMF, MI-ACE, and $e$FUMI estimated target concepts and EM-DD estimated scaled target concept.  MI-SMF and MI-ACE emphasize the distinguishing characteristics of the target concept which can be seen by the large value at 1$\mu$m and negative values at around 0.5$\mu$m which corresponds to wavelength regions in which the target concept has relatively large and small magnitude in comparison to background/non-target materials, respectively.  $e$FUMI attempts to reconstruct the true target signature shape which can be seen by comparing its target concept to the true target signature shown in Figure \ref{fig:constituent_endmembers}.  Finally, for EM-DD, the estimated target concepts scaled by the estimated feature scaling is shown.  The result illustrates that EM-DD is very effective at identifying the regions in which the target concept is different from background/non-target materials as a result of its estimated scaling parameters.  These target concepts were estimated using 50 positive and negative bags each containing ten data points.  Positive bags contained two true target points with an average target proportion of 0.9.  Zero-mean Gaussian noise was added to the data such that the SNR was 40dB.  }
\label{fig_sim2}
\end{figure}

Table \ref{tab:sim21} lists the AUC values for each experiment with results averaged over ten runs. MI-SMF and MI-ACE tended to outperform $e$FUMI and EM-DD in this experiment. As one would expect, results tend to improve over all methods as more true target points data points are available and degrade when only very few, highly mixed points are available (in this case, only 3 positive bags containing 2 true target points each resulting in a total of 6 target points with an average target proportion of 0.05).   However, even in the case of only a few positive bags, results improve when the average target portion or the number of target points per bag is increased.  For example, as a comparison, consider the case of 3 positive bags, containing 2 true target points each but the target proportion is increased to 0.25 on average, then average AUC($\pm$ std. dev.) over 10 runs improves to 0.995$\pm$0.001, 0.994$\pm$0.001, 0.840$\pm$0.167, 0.506$\pm$0.110 for MI-SMF, MI-ACE, $e$FUMI, and EM-DD, respectively.  Fig. \ref{fig_sim2} shows example estimated target concepts for each of the four methods. When examining the spectra estimated by MI-SMF and MI-ACE in Fig. \ref{fig_sim2} and comparing them to the true signatures used to generate the data in Fig. \ref{fig:constituent_endmembers}, the discriminative ability of the estimated target concepts become apparent and can be interpreted.  For example, observe that MI-SMF and MI-ACE estimate negative values around wavelength 0.5$\mu$m, when examining Fig. \ref{fig:constituent_endmembers} it can be seen that this corresponds to spectral wavelengths in which non-target signatures have relatively larger values as compared to the target, thus, the negative target signature in these wavelength impose a penalty for any large values at this wavelength in test data.  In contrast, large values of the estimated signatures around 1$\mu$m correspond to a wavelength region in which the target signature has a relatively large value in comparison to the background materials.  Finally, around 1.45$\mu$m the estimated values are close to zero indicating that the target signature and background materials have similar values at this wavelength.  

\begin{table} \centering
\caption{Simulated Hyperspectral Data Experiments: Varying Proportion of Positive vs. Negative Bags. The target proportion for each true target point was 0.05 on average in training data.  The average target proportion for each true target point in test data was 0.15. Results listed in Mean AUC $\pm$ Standard Deviation on separate Test Data over 10 Runs.  Best score in each experiment is in bold; Second best score in each experiment is underlined. }
\begin{tabular}{|l|c|c|c|L|} 
\hline
&  \multicolumn{3}{c|}{\textbf{Proportion of Positive Bags}}  & \multirow{2}{11mm}{\scriptsize{\textbf{Avg. Run Time (s)}}}\\\cline{2-4}

&\textbf{0.25}&\textbf{0.15}&\textbf{0.05} &  \\\hline
\hline
\textbf{MI-SMF} &\textbf{0.988$\pm$0.008}&\textbf{0.987$\pm$0.014}&\textbf{0.838$\pm$0.303}& 0.006\\\hline
\textbf{MI-ACE} &\underline{0.917$\pm$0.197}&\underline{0.979$\pm$0.030}&\underline{0.716$\pm$0.368}& 0.006\\\hline
\textbf{$e$FUMI}&0.907$\pm$0.054&0.818$\pm$0.234&0.651$\pm$0.346&1.082\\\hline
\textbf{EM-DD}   &0.549$\pm$0.117&0.568$\pm$0.148&0.476$\pm$0.087&0.822\\\hline
				\hline
\end{tabular}
\label{tab:sim21}
\end{table}

In this experiment,  some of the $e$FUMI results are on par with those of  MI-ACE.  This is expected as $e$FUMI is also a sub-pixel target characterization approach.  However, there are a number of significant advantages of MI-SMF and MI-ACE over $e$FUMI. One of these is running time. In general, MI-ACE and MI-SMF are faster than $e$FUMI (since $e$FUMI alternately computes sub-pixel proportion values for each data point and updates target and non-target concepts using a series of large matrix operations).  To show this, the average running time of our MATLAB implementations of MI-SMF, MI-ACE, $e$FUMI, and EM-DD (excluding initialization) for each simulated hyperspectral data experiment are listed alongside the results in Tables 1-3.  These experiments were run on a MacBookPro with 2.5Ghz quad-core Intel Core i7 processor and 16GB of RAM. Also, $e$FUMI attempts to recover the true target signature from the data (as opposed to a discriminative signature) and, thus, does not leverage contextual information when it may beneficial as discussed in the previous simulated data experiment. (However, in cases when the goal is to uncover the true target signature, $e$FUMI would be a better choice than MI-ACE or MI-SMF.) Furthermore, the resulting target signatures of MI-ACE and MI-SMF can be interpreted to determine which wavelengths are informative for the target detection problem and their relationship with respect to the background.  Large positive values in the resulting MI-ACE and MI-SMF target signature indicate that the target material has a larger response in those wavelengths when compared to the background.  Similarly, large negative values in the target signature indicate the target material has a smaller response in those wavelengths when compared to the background. Values close to zero indicate that the associated wavelength is not informative for the target detection problem.    Finally, $e$FUMI requires setting several parameters whereas MI-SMF and MI-ACE are parameter free.  Determining the many appropriate parameter settings for $e$FUMI (through cross validation) can often be time consuming.

\paragraph{Varying Number of Target Points in Positive Bags}
In this experiment, the number of target points in each positive bag was varied to be 25\%, 15\%, and 5\% of the points in the bag (corresponding to 3, 2, and 1 points, respectively).  The total number of bags was 50 and these were split evenly across positive and negative bags with each bag containing ten points total.  The target proportion for each true target point was 0.05 on average.  The resulting AUC values for each experiment averaged over 10 runs is shown in Table \ref{tab:sim22}.  As can be seen, the results in this experiment are similar to the previous one in that performance improves for MI-ACE, MI-SMF and $e$FUMI given more true target training data points. 
\begin{table} \centering
\caption{Simulated Hyperspectral Data Experiments: Varying Number of True Target Points in Positive Bags. The target proportion for each true target point was 0.05 on average in training data.  The average target proportion for each true target point in test data was 0.15.  Results listed in Mean AUC $\pm$ Standard Deviation on separate Test Data over 10 Runs.  }
\begin{tabular}{|l|c|c|c|L|}
\hline
&  \multicolumn{3}{c|}{\textbf{Proportion of Target Points in Positive Bags}} & \multirow{2}{11mm}{\scriptsize{\textbf{Avg. Run Time (s)}}}\\\cline{2-4}
&\textbf{0.25}&\textbf{0.15}&\textbf{0.05}&\\\hline
\hline
\textbf{MI-SMF}&\textbf{0.984$\pm$0.005}&\textbf{0.978$\pm$0.011}&\textbf{0.925$\pm$0.161}& 0.007\\\hline
\textbf{MI-ACE}&\underline{0.981$\pm$0.006}&         0.958$\pm$0.045&         0.811$\pm$0.224& 0.007\\\hline
\textbf{$e$FUMI}&       0.968$\pm$0.012&\underline{0.960$\pm$0.015}&\underline{0.906$\pm$0.074}& 1.240\\\hline
\textbf{EM-DD}&          0.485$\pm$0.049&         0.455$\pm$0.047&         0.494$\pm$0.045& 0.914\\\hline
				\hline
\end{tabular}
\label{tab:sim22}
\end{table}

\paragraph{Varying Target Proportion in True Target Points}
In the final simulated data experiment, the proportion of target in each true target point was varied to be 0.25, 0.15 and 0.05.   The total number of bags was 50 and these were evenly split across positive and negative bags with each bag containing ten points total.  The number of target points in each positive bag was set to two. Table \ref{tab:sim23} lists the AUC values for each experiment with results averaged over ten runs.  From these results, it can be seen that MI-SMF, MI-ACE and $e$FUMI are all effective with decreasing amounts of target proportion given enough true target data points in the training data (in this case, 50 true target points). 
\begin{table} \centering
\caption{Simulated Hyperspectral Data Experiments: Varying Mean Target Proportion in True Target Points. Results listed in Mean AUC $\pm$ Standard Deviation on separate Test Data over 10 Runs}
\begin{tabular}{|l|c|c|c|L|}
\hline
&  \multicolumn{3}{c|}{\textbf{Mean Target Proportion in True Target Points }} & \multirow{2}{11mm}{\scriptsize{\textbf{Avg. Run Time (s)}}}\\\cline{2-4}
&\textbf{0.25}&\textbf{0.15}&\textbf{0.05}&\\\hline
\hline
\textbf{MI-SMF}&       \textbf{0.989$\pm$0.002}&       \textbf{0.988$\pm$0.002}&    \textbf{0.984$\pm$0.003}& 0.008\\\hline
\textbf{MI-ACE}&    \underline{0.987$\pm$0.001}&    \underline{0.986$\pm$0.003}&   \underline{0.981$\pm$0.004}& 0.008\\\hline
\textbf{$e$FUMI}&             {0.985$\pm$0.002}&              {0.982$\pm$0.004}&            0.964$\pm$0.012& 1.04\\\hline
\textbf{EM-DD}&                 0.469$\pm$0.120 &               0.456$\pm$0.074&             0.486$\pm$0.106& 0.823\\\hline
				\hline
\end{tabular}
\label{tab:sim23}
\end{table}

\subsection{AR Face Sunglasses Detection}
Although the proposed approach was motivated by our work in sub-pixel hyperspectral target detection, the method may be applicable to a variety of other data types and applications.  Namely, MI-ACE/MI-SMF estimate discriminative target signatures from mixed and inaccurately labeled training data. The proposed method can be applied to any data set or application plagued with inaccurate training labels in which a target ``signature'' or linear discriminant is needed. To help illustrate this and to help better visualize the ability of MI-ACE and MI-SMF to identify discriminative features, MI-ACE and MI-SMF were also applied to an MIL detection problem constructed using the AR Face Data Set \cite{martinez:1998}.  The AR-face data set consists of frontal-pose images with 26 images/person (2 sessions, 13 per session) corresponding to different expressions, illuminations and occlusions. Pre-processed and cropped imagery of 50 male and 50 female subjects provided by Martinez and Kak \cite{martinez2001pca} was used. Each image was down-sampled to  $83\times 60$ pixels and the raw gray scale values were used as features.

For the experiment, \textit{sun-glasses} were selected as the {target concept}. Specifically, 50 positive training bags of 10 instances each were created.  Each positive bag contained only two instances of randomly selected images of people wearing sun-glasses; the other eight were randomly chosen from images of people without sun-glasses. 50 negative bags were constructed by randomly selecting 10 instances per bag of images of individuals not wearing sun-glasses. Test data included all imagery that was not used for training. Admittedly, the AR dataset is not naturally an MIL problem.  However, the purpose of these results is to simply illustrate that the approach can be effectively applied to other data types and to help the reader visualize the discriminative target signatures that are estimated by the MI-ACE and MI-SMF algorithms. 

MI-ACE and MI-SMF were applied to this data set along with the following comparison algorithms: $e$FUMI \cite{Jiao:2015}, EMDD (EM-DD in which the target point and scale is estimated), EMDD-P (EM-DD in which only a target point is estimated)\cite{Zhang:2002}, DMIL \cite{Shrivastava:2014,Shrivastava:2015} and mi-SVM \cite{andrews:2002}. The mi-SVM algorithm was added to these experiments to include a comparison MIL approach that does not rely on estimating a target signature. $e$FUMI was initialized as outlined in \cite{Jiao:2015} and run with parameter settings of $\mu = 0.05, \alpha = 1.2, \beta=60, M=7$, and $\Gamma = 10$. These $e$FUMI parameters were determined manually to maximize $e$FUMI performance.  EM-DD scaling values were all initialized to one and the target point value was initialized to the same initial positive data point as used by $e$FUMI.  Results are shown in Table \ref{tab:sim21b} and the target concept estimated by MI-ACE, MI-SMF, $e$FUMI, EMDD, EMDD-P, and DMIL are shown in Fig. \ref{fig_sim}. As can be seen by examining the table, MI-ACE, MI-SMF, and $e$FUMI outperform the other methods.  However, MI-ACE and MI-SMF have several significant advantages over $e$FUMI in obtaining these detection results.  Namely, MI-ACE and MI-SMF do not have parameters to set whereas $e$FUMI has a large number of parameters to tune. Furthermore, MI-ACE and MI-SMF have faster running time when compared to $e$FUMI.

\begin{table} \centering
\caption{Normalized Area Under the ROC Curve at False Alarm Rates of 0.001 and 1 for the AR Face Dataset ``Sunglasses'' MIL Detection experiment. }
\begin{tabular}{|c|c|c|} 
\hline
\multirow{2}{11mm}{\scriptsize{\textbf{Algorithm}}} & \multicolumn{2}{c|}{\textbf{NAUC}} \\\cline{2-3}
& FAR=0.001 & FAR=1 \\\hline
MI-SMF & \underline{0.998} & \textbf{1.000} \\\hline
MI-ACE & \textbf{1.000} & \textbf{1.000} \\\hline
$e$FUMI & \underline{0.998} & \textbf{1.000} \\\hline
EMDD & 0.210 & 0.772 \\\hline
EMDD-P & 0.776 & 0.987 \\\hline
DMIL & 0.798 & 0.991 \\\hline
mi-SVM & 0.671 & 0.989 \\\hline
\end{tabular}
\label{tab:sim21b}
\end{table}

\begin{figure}[!h]
\centering

\subfloat[MI-SMF estimated target concept ]{\includegraphics[width=2in]{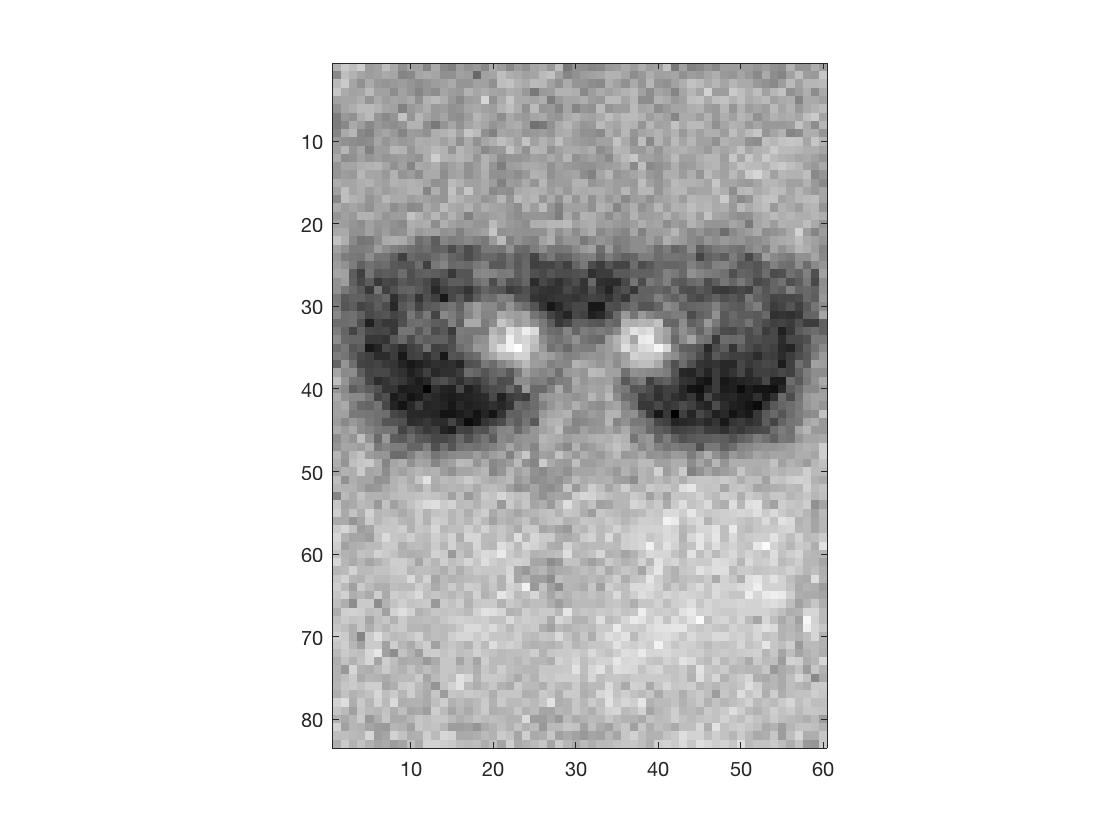}%
\label{fig:simexp1_2b}}
\subfloat[MI-ACE estimated target concept]{\includegraphics[width=2in]{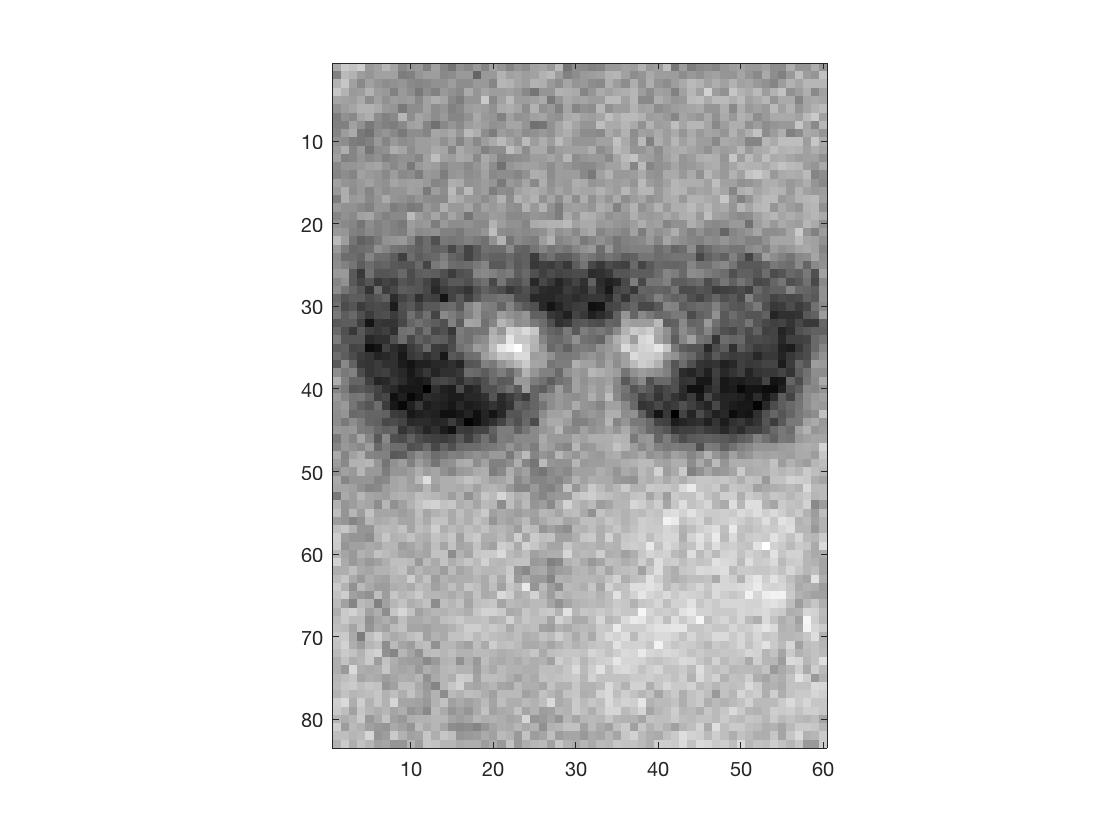}%
\label{fig:simexp1_3b}}

\subfloat[$e$FUMI estimated target concept]{\includegraphics[width=2in]{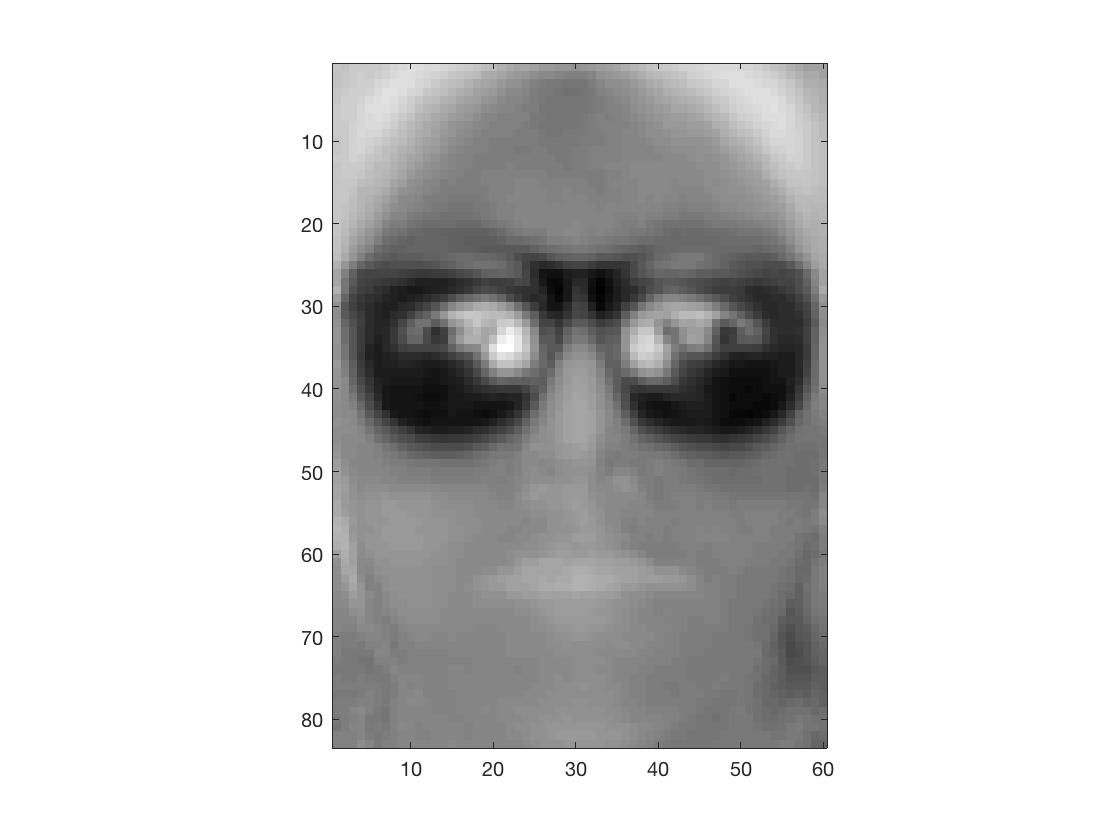}%
\label{fig:simexp1_4b}}
\subfloat[EMDD estimated target concept]{\includegraphics[width=2in]{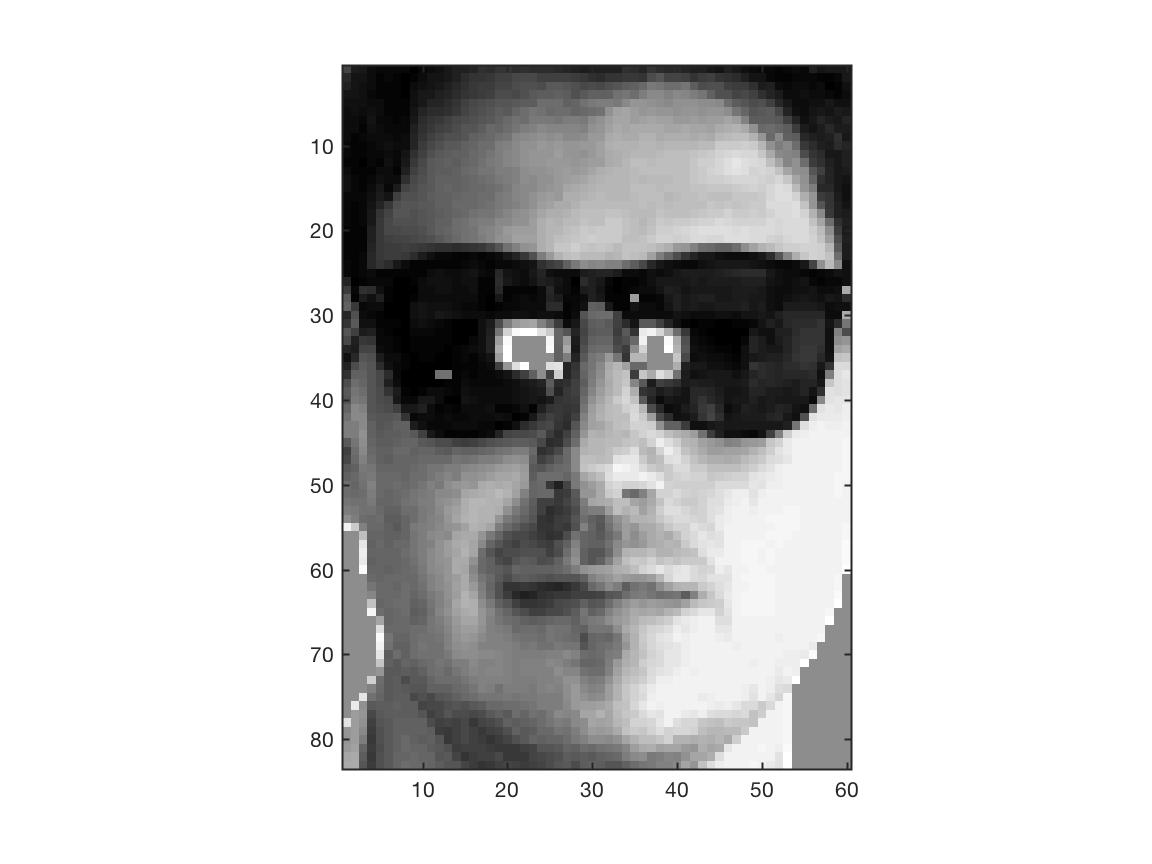}%
\label{fig:simexp1_5b}}

\subfloat[EMDD-P estimated target concept]{\includegraphics[width=2in]{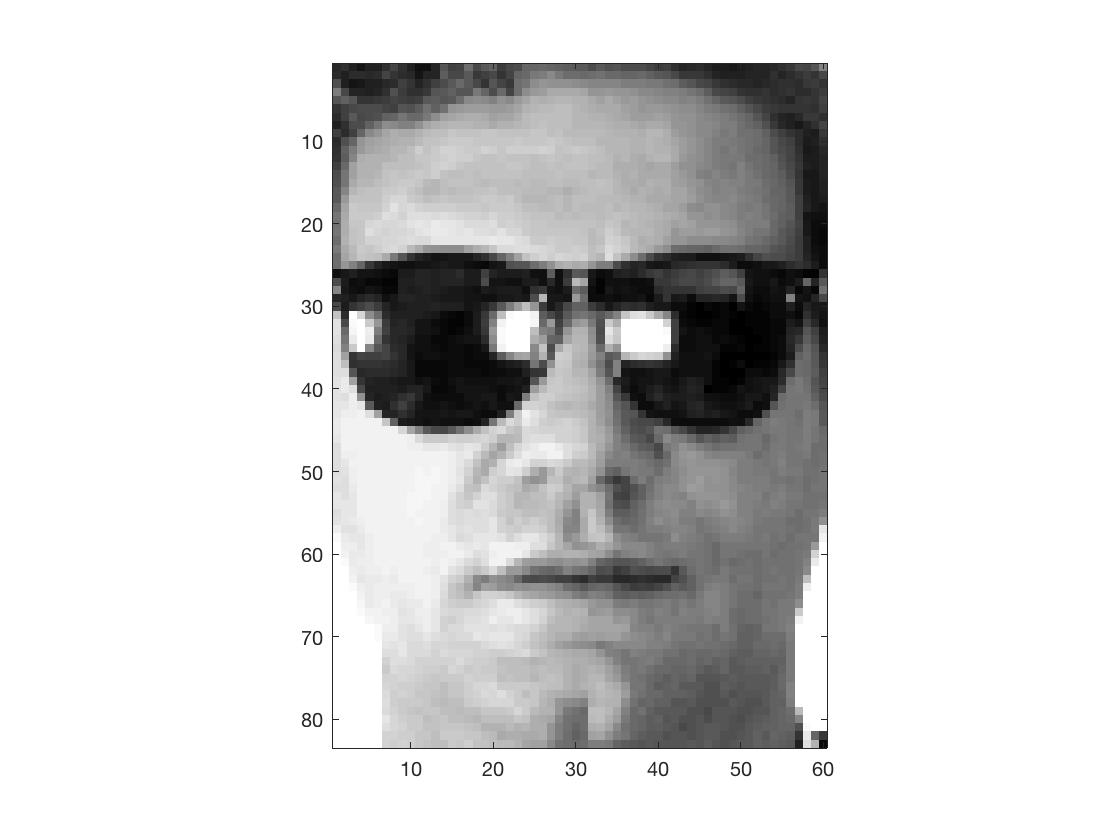}%
\label{fig:simexp1_6b}}
\subfloat[DMIL estimated target concept]{\includegraphics[width=2in]{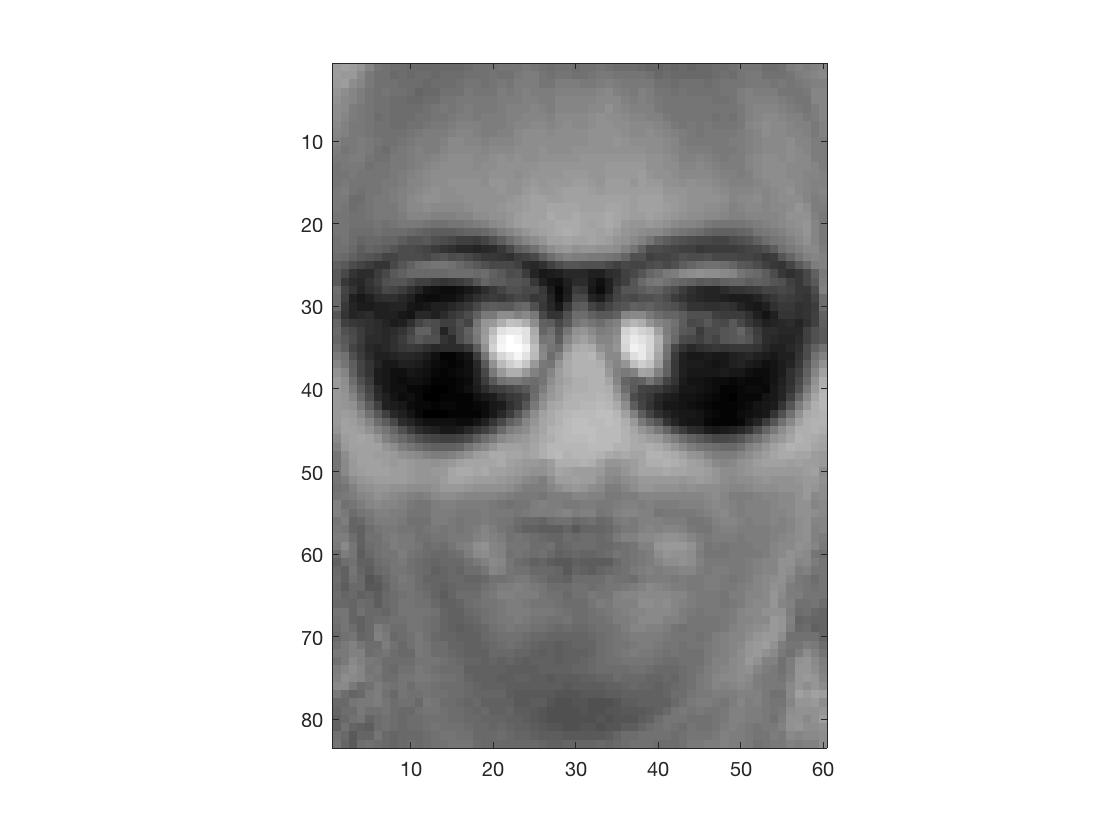}%
\label{fig:simexp1_4b}}

\caption{Target concepts estimated for the AR Face ``Sunglass'' Target Concept estimated using MI-SMF, MI-ACE, $e$FUMI, EMDD, EMDD-P and DMIL. As can be seen in these figures, MI-SMF and MI-ACE correctly highlight sunglasses as the most discriminative features in the imagery to distinguish between the target and non-target data points. }
\label{fig_sim}
\end{figure}

\subsection{MUUFL Gulfport Hyperspectral Data}
\begin{figure}[h]
\centering
{\includegraphics[width=3.5in]{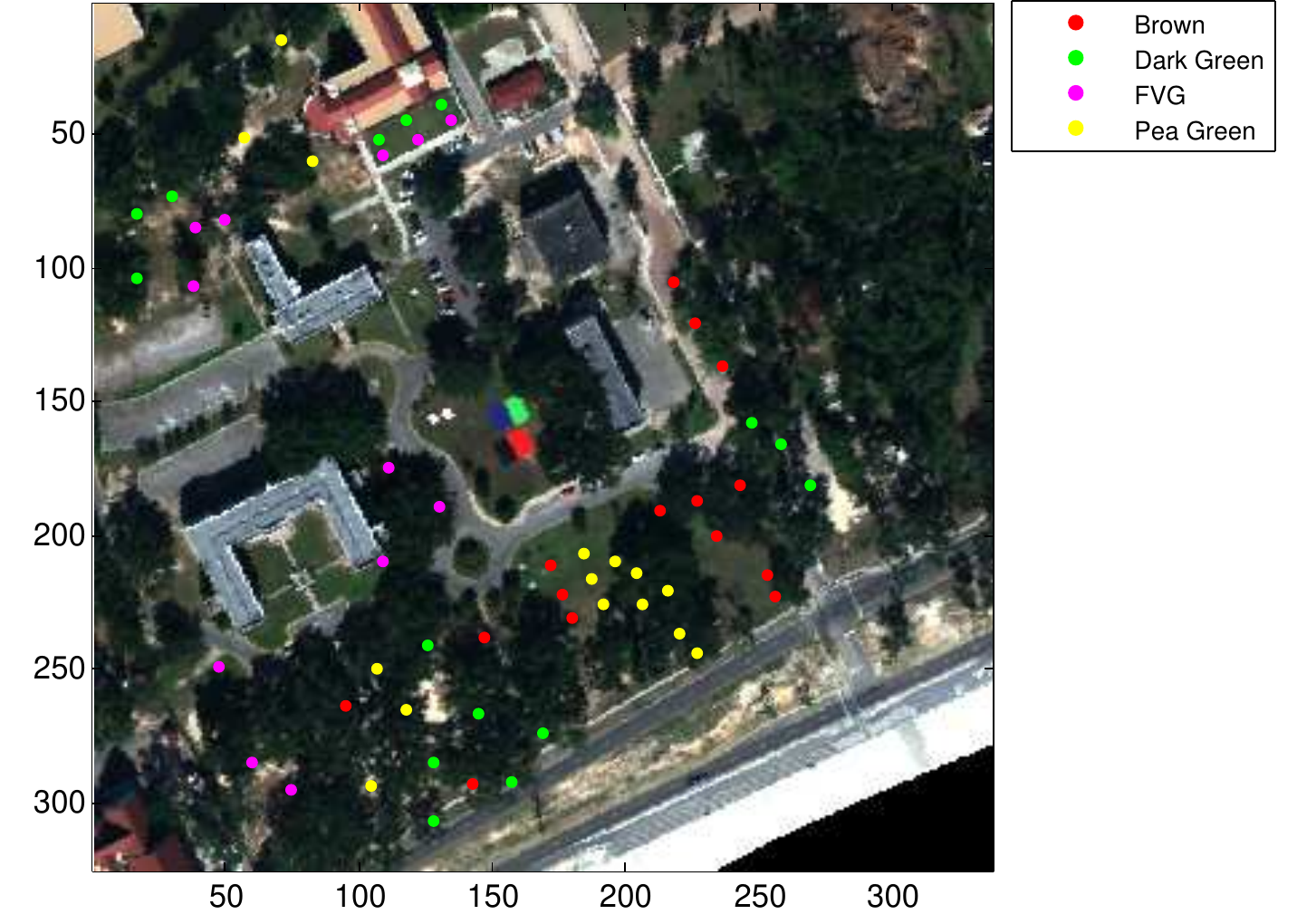}}
\caption{MUUFL Gulfport data set RGB image and the 64 target locations}
\label{fig:gulfport_rgb}
\end{figure}

For experiments on real hyperspectral target detection data, the MUUFL Gulfport Hyperspectral data set was used.  This data set was collected over the University of Southern Mississippi-Gulfpark Campus and contains $325\times337$ pixels with 72 bands corresponding to wavelengths from $367.7 nm$ to $1043.4 nm$ at a $9.5-9.6 nm$ spectral sampling interval \cite{gader:2013}. The first four and last four spectral bands were removed from the data set due to noise.  The spatial resolution is 1 m. Two flights over the area from this data (Gulfport Campus Flight 1 and Gulfport Campus Flight 3) were selected as cross-validated training and testing data. These flights were selected as they were flown at the same altitude and have the same spatial resolution.  Throughout the scene, there are 57 emplaced man-made targets.  The targets are cloth panels of four different colors: Brown (15 examples), Dark Green (15 examples), Faux Vineyard Green (FVG) (12 examples) and Pea Green (15 examples).  The spatial location of the targets are shown as scattered points over an RGB image of the scene in Fig. \ref{fig:gulfport_rgb}. This data set is a very challenging target detection task as many of the targets are partially or fully occluded by Live Oak trees on the campus.  Furthermore, the targets  vary in size, for each target type, there are targets that are $0.25 m^2$, $1 m^2$ and $9 m^2$ in area.  Thus, a target that has $0.25 m^2$ covers at most (in the case when it is fully within the footprint of a single pixel) a 0.25 proportion of the pixel signature.  Many of these targets straddle multiple pixels and are occluded resulting in a highly mixed, sub-pixel target detection task.  For each target in the training flight, a $5\times5$ rectangular region around each ground truth point for each target were labeled as positive bags; this size was chosen since the accuracy of the GPS device used to record the groundtruth locations had 5m accuracy. Thus, there are 57 positive bags in each training set in this experiment. 

MI-SMF and MI-ACE were evaluated on this data using the Normalized Area Under the receiver operating characteristic Curve (NAUC) in which the area was normalized out to a false alarm rate (FAR) of 1$\times 10^{-3}$ false alarms$/m^2$  \cite{glenn_gulfport:2013}.  Given the spatial resolution of the imagery, this maximum FAR corresponds to one false alarm per 1000 pixels.  An NAUC value of one corresponds to zero false alarms and 100\% detection.   MI-ACE and MI-SMF were compared to the $e$FUMI \cite{Jiao:2015}, EM-DD, EM-DD-P\cite{Zhang:2002}, mi-SVM \cite{andrews:2002}, and DMIL \cite{Shrivastava:2014,Shrivastava:2015} algorithms.    For all methods except mi-SVM and EM-DD, target concepts were estimated on the training flight and then used to perform detection on the test flight using the ACE detection statistic.  During application of ACE on the test data, the background mean and covariance were estimated from the negative instances of the training data.   Since mi-SVM does not estimate a target concept, the detection statistic used for the mi-SVM approach was the signed distance to the decision hyperplane estimated on the training data. For EM-DD, in order to effectively make use of the scale parameters learned, the detection statistic in \eqref{eqn:EM-DDprd} was used as outlined in \cite{Zhang:2002}.   Given an initialization, all methods obtained consistent results when re-run except for $e$FUMI, EM-DD and EM-DD-P whose initialization procedures include a stochastic step.  Thus, the results reported for $e$FUMI, EM-DD and EM-DD-P are the median results over five runs of the algorithm on the same data.  
\subsubsection{Single Negative Bag}

In the first MUUFL Gulfport experiment,  one negative bag composed of all instances in the training data outside of any positive bag was used during training.   The results of MI-SMF, MI-ACE and comparison methods are shown in Table \ref{tab:gulf1}.  As can be seen, MI-SMF and/or MI-ACE provide consistently either the best or second best result in comparison to the other approaches. 

\begin{table*}[!htb]
		\begin{center}
			\vspace{-4mm}\caption{NAUC on MUUFL Gulfport, One Negative Bag}  \label{tab:gulf1}
			\begin{tabular}{|c|c|c|c|c|c|c|c|c|}
				\hline
				\multirow{2}{*}{Alg.} &  \multicolumn{4}{c|} {Train on Flight 1; Test on Flight 3 } &  \multicolumn{4}{c|} {Train on Flight 3; Test on Flight 1 } \\
				\cline{2-9}      &      Brown    &    Dark Gr.  &  Faux Vine Gr. &   Pea Gr.    &    Brown      &    Dark Gr.   & Faux Vine Gr.  &   Pea Gr.   \\    
				\hline\hline
				\textbf{MI-SMF}  &          {0.448}&\underline{0.382}&\underline{0.579}&\underline{0.316}&   \textbf{0.760}&\underline{0.501}&   \textbf{0.650}&\underline{0.384}    \\\hline
				\textbf{MI-ACE}  &   \textbf{0.474}&   \textbf{0.390}&           0.485 &   \textbf{0.333}&   \textbf{0.760}&           0.483 &\underline{0.593}&           0.380     \\\hline
				\textbf{$e$FUMI} &          0.433  &           0.377 &   \textbf{0.707}&           0.267 &\underline{0.753}&   \textbf{0.502}&           0.470 &   \textbf{0.394}    \\\hline
				\textbf{mi-SVM}  &          0.353  &           0.265 &           0.437 &           0.265 &           0.333 &           0.368 &           0.243 &           0.268     \\\hline
				\textbf{EM-DD-P} &\underline{0.467}&           0.0   &           0.067 &           0.014 &           0.0   &           0.0   &           0.291 &           0.0       \\\hline
				\textbf{EM-DD} &            {0.0}&           0.0   &           0.055 &           0.0   &           0.0   &           0.0   &           0.0   &           0.0       \\\hline
				\textbf{DMIL}    &          0.418  &\underline{0.382}&           0.288 &           0.021 &           0.751 &           0.310 &           0.083 &           0.111     \\\hline
			\end{tabular}
		\end{center}
\end{table*}

\begin{figure*}[htbp] 
	\centering 
	\subfloat[Target Concepts Estimated for Brown]{
		\includegraphics[trim={2.3cm 6.6cm 2.2cm 7cm},clip,width=6.8cm]{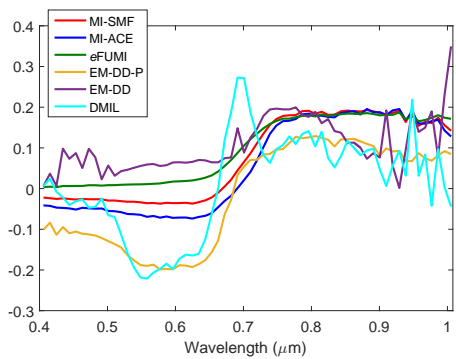} \label{fig:gulfport_sigs_brown}
	}
	\subfloat[Target Concepts Estimated for Dark Green ]{
		\includegraphics[trim={2.3cm 6.6cm 2.2cm 7cm},clip,width=6.8cm]{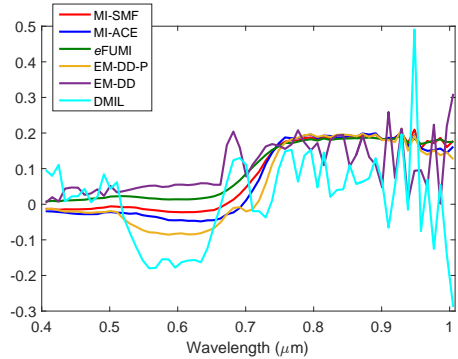} \label{fig:gulfport_sigs_dgreen}
	} 

	\subfloat[Target Concepts Estimated for Faux Vineyard Green ]{
		\includegraphics[trim={2.3cm 6.6cm 2.2cm 7cm},clip,width=6.8cm]{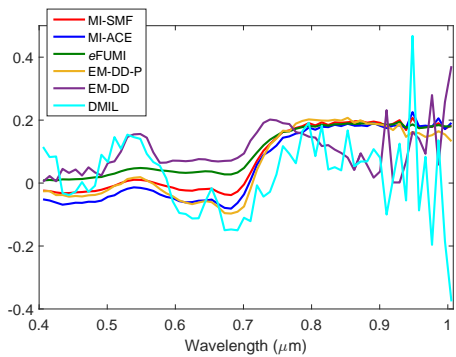} \label{fig:gulfport_sigs_fvgreen}
	}
	\subfloat[Target Concepts Estimated for Pea Green ]{
		\includegraphics[trim={2.3cm 6.6cm 2.2cm 7cm},clip,width=6.8cm]{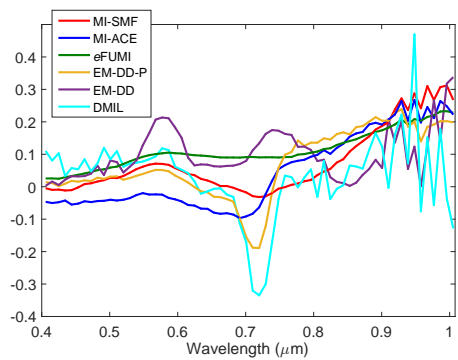} \label{fig:gulfport_sigs_pgreen}
	}

	\caption{Comparison of target concepts found by MI-SMF, MI-ACE, $e$FUMI, EM-DD-P, EM-DD and DMIL for (a) Brown, (b) Dark Green, (c) Faux Vineyard Green, and (d) Pea Green Targets} 
	\label{fig:gulfport_estimated_sigs}
\end{figure*}

Fig. \ref{fig:gulfport_estimated_sigs} shows the target concepts estimated by MI-SMF, MI-ACE, $e$FUMI, mi-SVM, EM-DD-P and DMIL. As can be seen, MI-SMF, MI-ACE and $e$FUMI tend to find target concepts with similar spectral shape; this agrees with the performance reported in Tab. \ref{tab:gulf1} in which MI-SMF, MI-ACE and $e$FUMI tend to have the best performance in this experiment.  One observation that can be made is that the results when tested on Flight 1 outperform those when tested on Flight 3.  This is due to the challenging nature of this detection problem.  As stated above, the targets are heavily mixed sub-pixel targets and there is heavy occlusion from tree coverage throughout the scene.  Although the flights cover the same general area, they do not have identical flight paths.  Thus, there are differences in viewing angle between the flights, differences in the field of view associated with each pixel, and, also, there may have been movement in tree branches.  These factors would result in differences in the number of sub-pixel targets that are visible across each flight.  In this case, testing on Flight 3 is more challenging.  This is seen consistently across many experiments and methods.  Similarly, across many experiments and methods, Dark Green and Pea Green targets tend to have lower detection rates (likely due to these targets being more occluded).

Both Flight 1 and Flight 3 were collected on the same day shortly after one another and, thus, were collected under similar environmental conditions.  The spectral signatures of materials vary across differing environmental conditions \cite{zare2014endmember}.  Since the MI-ACE and MI-SMF algorithms learn a discriminative target signatures that distinguishes the target spectral signature from background material, the performance of MI-ACE and MI-SMF depends upon the magnitude and spectral shape of the target vs. background materials to maintain the same relative relationship.  In other words, if we train MI-ACE or MI-SMF on data collected in one set of environmental conditions and test on data collected in different environmental conditions, the performance of the methods will depend on whether the relative magnitudes of the target and background materials are similar to each other across the environmental conditions.  If, for example, MI-ACE placed a large positive weight on a band since the target material has a large spectral response as compared to the background in that wavelength, MI-ACE would perform well on the test data if the target material still had a comparatively large spectral response in that wavelength.  However, results would degrade if the relative values of the target and background materials were swapped.  As with all supervised learning methods, the ability of the approach to generalize to test data is dependent on how well the training data distribution matches or encompasses what is seen during test.

\subsubsection{Multiple Negative Bags}
In the second MUUFL Gulfport experiment, the influence of the construction of the negative bags was examined. In the previous experiment, only one negative bag consisting of all instances outside of any positive bag was used.  Using one negative bag in MI-SMF and MI-ACE results in each instance in the negative bag to have equal influence on the result.  Thus, if one or a few background/non-target materials compose the majority of the instances in the negative bag, these materials have a larger impact and influence on the estimated target concept.  In the case of EM-DD and EM-DD-P, negative bag construction influences results heavily as, for these approaches, a single instance from each bag is used to represent the bag during target concept and scale updates.  Thus, given only one negative bag, only one negative instance is used to represent all of the negative data.     In this experiment, we investigate the use of multiple negative bags.  To construct the multiple negative bags, all instances outside of any positive bag are clustered using the $K$-means clustering algorithm and each resulting cluster is used as a separate negative bag.  The purpose of this approach is to cluster together the instances with similar spectral shape and magnitude.  When running the $K$-means algorithm, $K$ was first set to 15 such that the number of negative bags is equal to the number of positive bags for most of the target types.  $K$ was then varied to be 100 and equal to the number of non-target instances in the data (i.e., each instance is an individual negative bag).   Table \ref{tab:kmeangulf}, \ref{tab:kmeangulf2}, and \ref{tab:kmeangulf3} list the results of MI-SMF, MI-ACE and comparison methods for $K = 15, 100$, and $106548$, respectively.  When studying Tab. \ref{tab:kmeangulf} - \ref{tab:kmeangulf3}, it can be seen that MI-SMF and MI-ACE are fairly consistent in their results, thus, MI-SMF and MI-ACE are not heavily influenced by negative bag structure.  The $e$FUMI, mi-SVM, and DMIL methods are also not influenced by negative bag structure and results are similar or the same as those with one large negative bag.  However, EM-DD-P is heavily influenced by the number of negative bags.  Results for EM-DD-P improve as more negative bags are included with the best results provided when each non-target point is an individual negative bag.  However, even with a negative bag for each non-target instance, MI-SMF and MI-ACE provide competitive results with EM-DD-P. 

\section{Summary}
MI-SMF and MI-ACE, two multiple instance target characterization approaches, are introduced as methods to estimate hyperspectral target signatures from imprecisely labeled training data.  Advantages of MI-SMF and MI-ACE include that they have a straight-forward implementation, fast running time,  and are free of parameter settings. Experimental results show that MI-SMF and MI-ACE provide competitive and state-of-the-art results when compared to existing multiple instance concept learning methods. Although this work was motivated by sub-pixel hyperspectral target detection, the MI-ACE and MI-SMF methods are general approaches for extracting discriminative target signatures given high dimensional data points (or feature vectors) that are paired with uncertain training labels.

\begin{table*}[!htb]
		\begin{center}
			\vspace{-4mm}\caption{NAUC on MUUFL Gulfport, Cluster Background into K=15 Negative Bags}\label{tab:kmeangulf}
			\begin{tabular}{|c|c|c|c|c|c|c|c|c|}
				\hline
        \multirow{2}{*}{Alg.} &  \multicolumn{4}{c|} {Train on Flight 1; Test on Flight 3 } &  \multicolumn{4}{c|} {Train on Flight 3; Test on Flight 1 } \\
				\cline{2-9}      &          Brown    &        Dark Gr.  &     Faux Vine Gr. &     Pea Gr.    &       Brown      &      Dark Gr.     & Faux Vine Gr.  &     Pea Gr.   \\    
				\hline\hline
				\textbf{MI-SMF}  &\underline{0.461}&\underline{0.382}&\underline{0.540}&\underline{0.320}&   \textbf{0.763}&   \textbf{0.503}&   \textbf{0.651}&\underline{0.374}    \\\hline
				\textbf{MI-ACE}  &   \textbf{0.496}&   \textbf{0.389}&           0.479 &   \textbf{0.333}&   \textbf{0.763}&           0.486 &\underline{0.565}&           0.349     \\\hline
				\textbf{$e$FUMI} &          0.433  &           0.377 &   \textbf{0.707}&           0.267 &\underline{0.753}&\underline{0.502}&           0.470 &   \textbf{0.394}    \\\hline
				\textbf{mi-SVM}  &          0.353  &           0.265 &           0.437 &           0.265 &           0.333 &           0.368 &           0.243 &           0.268     \\\hline
				\textbf{EM-DD-P} &          0.038  &           0.0   &           0.0   &           0.0   &           0.284 &           0.012 &           0.130 &           0.019     \\\hline
				\textbf{EM-DD} &          0.0    &           0.0   &           0.0   &           0.086 &           0.0   &           0.0   &           0.0   &           0.0       \\\hline
				\textbf{DMIL}    &          0.418  &\underline{0.382}&           0.288 &           0.021 &           0.751 &           0.310 &           0.083 &           0.111     \\\hline
			\end{tabular}
		\end{center}
\end{table*}

\begin{table*}[!htb]
		\begin{center}
			\vspace{-4mm}\caption{NAUC on MUUFL Gulfport, Cluster Background into K=100 Negative Bags}\label{tab:kmeangulf2}
			\begin{tabular}{|c|c|c|c|c|c|c|c|c|}
				\hline
        \multirow{2}{*}{Alg.} &  \multicolumn{4}{c|} {Train on Flight 1; Test on Flight 3 } &  \multicolumn{4}{c|} {Train on Flight 3; Test on Flight 1 } \\
				\cline{2-9}      &          Brown    &        Dark Gr.  &     Faux Vine Gr. &     Pea Gr.    &       Brown      &      Dark Gr.     & Faux Vine Gr.  &     Pea Gr.   \\    
				\hline\hline
				\textbf{MI-SMF}  &\underline{0.454}&          {0.382}&\underline{0.560}&\underline{0.312}&   \textbf{0.762}&   \textbf{0.506}&   \textbf{0.651}&\underline{0.379}    \\\hline
				\textbf{MI-ACE}  &   \textbf{0.476}&   \textbf{0.389}&           0.484 &   \textbf{0.333}&   \textbf{0.762}&           0.486 &\underline{0.558}&           0.366     \\\hline
				\textbf{$e$FUMI} &          0.433  &           0.377 &   \textbf{0.707}&           0.267 &\underline{0.753}&\underline{0.502}&           0.470 &   \textbf{0.394}    \\\hline
				\textbf{mi-SVM}  &          0.353  &           0.265 &           0.437 &           0.265 &           0.333 &           0.368 &           0.243 &           0.268     \\\hline
				\textbf{EM-DD-P} &          0.122  &\underline{0.386}&           0.0   &           0.267 &           0.066 &           0.013 &           0.545 &           0.265     \\\hline
				\textbf{EM-DD} &          0.001 &            0.0   &           0.0   &           0.020 &           0.046 &           0.0   &           0.0   &           0.026     \\\hline
				\textbf{DMIL}    &          0.418  &          {0.382}&           0.288 &           0.021 &           0.751 &           0.310 &           0.083 &           0.111     \\\hline
			\end{tabular}
		\end{center}
\end{table*}

\begin{table*}[!htb]
		\begin{center}
			\vspace{-4mm}\caption{NAUC on MUUFL Gulfport, Each Background Point is an Individual Negative Bag, K=106548}\label{tab:kmeangulf3}
			\begin{tabular}{|c|c|c|c|c|c|c|c|c|}
				\hline
        \multirow{2}{*}{Alg.} &  \multicolumn{4}{c|} {Train on Flight 1; Test on Flight 3 } &  \multicolumn{4}{c|} {Train on Flight 3; Test on Flight 1 } \\
				\cline{2-9}      &      Brown    &    Dark Gr.  &  Faux Vine Gr. &   Pea Gr.    &    Brown      &    Dark Gr.   & Faux Vine Gr.  &   Pea Gr.   \\    
				\hline\hline
				\textbf{MI-SMF}  &\underline{0.448}&\underline{0.382}&\underline{0.579}&\underline{0.316}&   \textbf{0.760}&          {0.501}&   \textbf{0.650}&          {0.384}    \\\hline
				\textbf{MI-ACE}  &   \textbf{0.474}&   \textbf{0.390}&           0.485 &   \textbf{0.333}&   \textbf{0.760}&           0.483 &\underline{0.593}&           0.380     \\\hline
				\textbf{$e$FUMI} &          0.433  &           0.377 &   \textbf{0.707}&           0.267 &          {0.753}&\underline{0.502}&           0.470 &\underline{0.394}    \\\hline
				\textbf{mi-SVM}  &          0.353  &           0.265 &           0.437 &           0.265 &           0.333 &           0.368 &           0.243 &           0.268     \\\hline
				\textbf{EM-DD-P} &          {0.420}&\underline{0.382}&           0.478 &           0.250 &\underline{0.759}&   \textbf{0.507}&           0.533 &   \textbf{0.422}     \\\hline
				\textbf{EM-DD} &            {0.0}&           0.064 &           0.0   &           0.0   &           0.0   &           0.0   &           0.055 &           0.0       \\\hline
				\textbf{DMIL}    &          0.418  &\underline{0.382}&           0.288 &           0.021 &           0.751 &           0.310 &           0.083 &           0.111     \\\hline
			\end{tabular}
		\end{center}
\end{table*}

\appendix[Derivation of Target Signature Updates]

%
In order to derive the update equation for the target signature of MI-ACE and MI-SMF, we can write the Lagrangian for following MI-ACE objective function shown in \eqref{eqn:objective2} as shown below:
\begin{equation}
\mathscr{L} = \frac{1}{N^+} \sum_{j: L_j = 1} \doublehat{\mathbf{s}}^T\doublehat{\mathbf{x}}^{\ast}_j - \frac{1}{N^-}\sum_{j:L_j = 0}\sum_{\mathbf{x}_i \in B_j^-} \doublehat{\mathbf{s}}^T\doublehat{\mathbf{x}_i} - \lambda\left(\doublehat{\mathbf{s}}^T\doublehat{\mathbf{s}} - 1\right)
\end{equation}
where $\lambda$ is the Lagrange multiplier.  The derivative of the Lagrangian with respect to $\doublehat{\mathbf{s}}$ is:
\begin{equation}
\frac{\partial \mathscr{L}}{\partial \doublehat{\mathbf{s}}} = \frac{1}{N^+} \sum_{j: L_j = 1} \doublehat{\mathbf{x}}^{\ast}_j - \frac{1}{N^-}\sum_{j:L_j = 0}\sum_{\mathbf{x}_i \in B_j^-} \doublehat{\mathbf{x}_i} - 2\lambda\doublehat{\mathbf{s}}
\label{eqn:lder}
\end{equation}
We can then set \eqref{eqn:lder} to zero and solve for $\doublehat{\mathbf{s}}$:
\begin{equation}
\doublehat{\mathbf{s}} = \frac{1}{2\lambda}\left(\frac{1}{N^+} \sum_{j: L_j = 1} \doublehat{\mathbf{x}}^{\ast}_j - \frac{1}{N^-}\sum_{j:L_j = 0}\sum_{\mathbf{x}_i \in B_j^-} \doublehat{\mathbf{x}_i} \right)
\end{equation}
Then, define $\mathbf{t}$ as:
\begin{equation}
\mathbf{t} = \frac{1}{N^+} \sum_{j: L_j = 1} \doublehat{\mathbf{x}}^{\ast}_j - \frac{1}{N^-}\sum_{j:L_j = 0}\sum_{\mathbf{x}_i \in B_j^-} \doublehat{\mathbf{x}_i}.
\end{equation}
To determine the value of the Lagrange multiplier, $\lambda$, we must determine the value for $\lambda$ that enforces the constraint that $\doublehat{\mathbf{s}}^T\doublehat{\mathbf{s}} = 1$.  Thus, $\lambda = \frac{\left\|\mathbf{t}\right\|}{2}$ which results in the final update equation for $\doublehat{\mathbf{s}}$:
\begin{equation}
\doublehat{\mathbf{s}} = \frac{\mathbf{t}}{\left\|\mathbf{t}\right\|} \text{ where } \mathbf{t} = \frac{1}{N^+} \sum_{j: L_j = 1} \doublehat{\mathbf{x}}^{\ast}_j - \frac{1}{N^-}\sum_{j:L_j = 0}\frac{1}{N_j^-}\sum_{\mathbf{x}_i \in B_j^-} \doublehat{\mathbf{x}_i}.
\end{equation}
The derivation for the update equation for the  MI-SMF target signature is identical to what is shown above except $\hat{\mathbf{x}}$ is used in place of $\doublehat{\mathbf{x}}$ in all of the preceding equations in this Section.

\section*{Acknowledgments}
This material is based upon work supported by the National Science Foundation under Grant No. IIS-1350078 - CAREER: Supervised Learning for Incomplete and Uncertain Data. The authors would also like to acknowledge James Theiler and Amanda Ziemann for their insightful discussions.

\ifCLASSOPTIONcaptionsoff
  \newpage
\fi



%
\bibliographystyle{IEEEtran}
\bibliography{FUMI_Ref}

%

\begin{IEEEbiography}[{\includegraphics[width=1in,height=1.25in,clip,keepaspectratio]{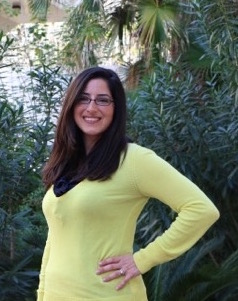}}]{Alina Zare}
received the Ph.D. degree from the University of Florida, Gainesville, in 2008. She is currently an Associate Professor with the Department of Electrical and Computer Engineering, University of Florida. Her research interests include machine learning and pattern recognition, sparsity promotion, target detection, hyperspectral image analysis, and remote sensing. Alina Zare is a recipient of the 2014 National Science Foundation CAREER award and the 2014 National Geospatial-Intelligence
Agency New Investigator Program Award.
\end{IEEEbiography}

\begin{IEEEbiography}[{\includegraphics[width=1in,height=1.25in,clip,keepaspectratio]{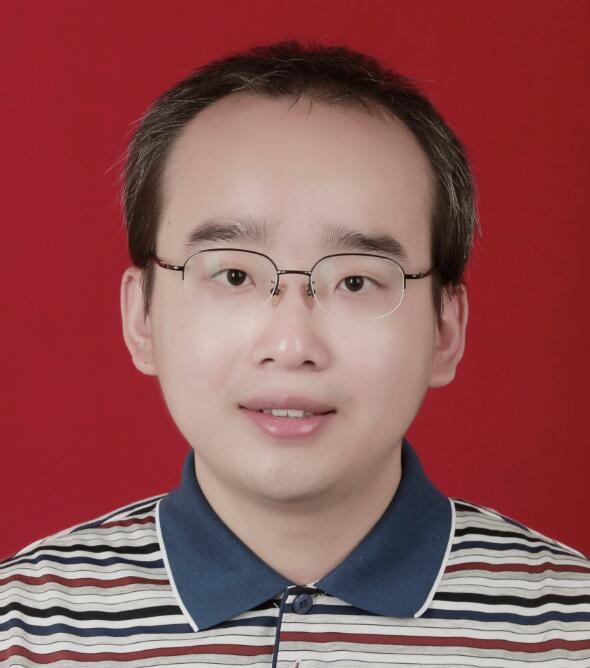}}]{Changzhe Jiao}
received the B.S. and M.S. degree in control theory both from Xidian University, Xi’an, China, in 2007 and 2012, respectively. He is currently a Graduate Research Assistant working toward the Ph.D. degree with the Department of Electrical Engineering and Computer Science, University of Missouri, Columbia, MO USA. His research interests include pattern recognition, multiple instance learning and hyperspectral image analysis.
\end{IEEEbiography}


\begin{IEEEbiography}[{\includegraphics[width=1in,height=1.25in,clip,keepaspectratio]{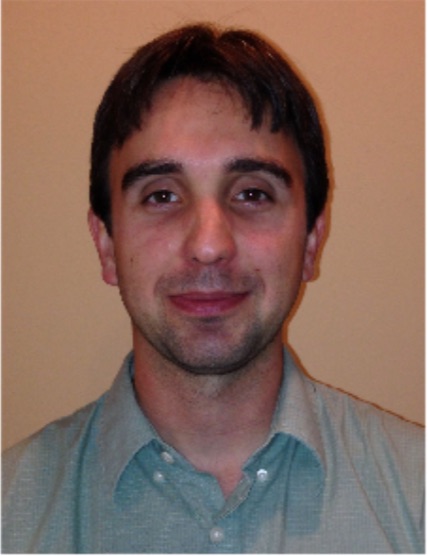}}]{Taylor Glenn}
received the Ph.D. degree from the University of Florida, Gainesville, FL, USA, in December 2013.
His research work has specialized in pattern recognition and machine learning and the application of these methods to sensor data. Before receiving his graduate degree, he cofounded 2G Engineering LLC, a company specializing in embedded systems design and quick-turn manufacturing. He has since founded Precision Silver LLC, a company producing image and data analysis software with a focus on aerial imagery and precision agriculture applications.
\end{IEEEbiography}




\end{document}